# An Algebra of Lightweight Ontologies

Marco A. Casanova, Rômulo Magalhães

Department of Informatics – PUC-Rio – Rio de Janeiro, RJ – Brazil
casanova@inf.puc-rio.br

**Abstract**. This paper argues that certain ontology design problems are profitably addressed by treating ontologies as theories and by defining a set of operations that create new ontologies, including their constraints, out of other ontologies. The paper first shows how to use the operations in the context of ontology reuse, how to take advantage of the operations to compare different ontologies, or different versions of an ontology, and how the operations may help design mediated schemas in a bottom up fashion. The core of the paper discusses how to compute the operations for lightweight ontologies and addresses the question of minimizing the set of constraints of a lightweight ontology. Finally, the paper describes an implementation of the operations, as a Protégé plug-in.

## 1 Introduction

In this paper, we argue that certain familiar ontology design problems are profitably addressed by treating ontologies as theories and by defining a set of operations that create new ontologies, including their constraints, out of other ontologies. We show how to compute the operations for *lightweight ontologies*, that is, ontologies whose constraints are *lightweight inclusions*. This class of constraints is expressive enough to cover the types of constraints commonly used in conceptual modeling and is as expressive as the class of inclusions considered in DL-Lite core with arbitrary number restrictions (Artale et al., 2009). We also discuss how to minimize the set of lightweight inclusions.

In more detail, we define an *ontology* as a pair $O=(V,\Sigma)$ such that $V$ is a *vocabulary* and $\Sigma$ is a set of *constraints* in $V$. The *theory of* $\Sigma$ is the set of all constraints that are logical consequences of $\Sigma$. We emphasize that the constraints in $\Sigma$ capture the semantics of the terms in $V$ and must, therefore, be brought to the foreground. The theory of $\Sigma$ in turn identifies the constraints that are implicitly defined, but which must be considered when using fragments of the ontology.

Turning to the ontology design problems that motivated the introduction of the operations, consider first the problem of designing an ontology to publish data on the Web. If the designer follows the Linked Data principles (Bernes-Lee, 2006; Bizer et al., 2007), he must select known ontologies, as much as possible, to organize the data so that applications "can dereference the URIs that identify vocabulary terms in order to find their definition". We argue that the designer should go further and analyze the constraints of the ontologies from which he is drawing the terms to construct his vocabulary. Furthermore, he should publish the data in such a way that the original semantics of the terms is preserved. To facilitate ontology design from this perspective, we introduce three operations on ontologies, called *projection*, *union* and *deprecation*.

Consider now the problem of comparing the expressive power of two ontologies, $O_1=(V_1,\Sigma_1)$ and $O_2=(V_2,\Sigma_2)$. If the designer wants to know what they have in common, he should create a mapping between their vocabularies and detect which constraints hold in both ontologies, after the terms are appropriately mapped. The *intersection* operation answers this question. We argued elsewhere (Casanova et al., 2010) that intersection is also useful to address the design of mediated schemas that combine several export schemas in a way that the data exposed by the mediator is always consistent.

On the other hand, if the designer wants to know what holds in $O_1=(V_1,\Sigma_1)$, but not in $O_2=(V_2,\Sigma_2)$, he should again create a mapping between their vocabularies and detect which constraints hold in the theory of $\Sigma_1$, but not in the theory of $\Sigma_2$, after the terms are appropriately mapped. The *difference* operation answers this question.

Finally, a variant of ontology comparison is the problem of analyzing what changed from one version of an ontology to the other. Difference is especially useful here.

To compute the operations, we adopt the machinery developed in (Casanova et al., 2010, 2012b) to handle lightweight inclusions. Previous work by the authors (Casanova et al., 2011; Casanova et al., 2012a) introduced the notion of *open fragment*, which is captured by the projection operation, and some of the operations discussed in detail in this paper. An early implementation of the operations is described in (Pinheiro, 2013).

The paper is organized as follows. Section 2 presents the formal definition of the operations. Section 3 illustrates the use of the operations to address ontology design problems. Section 4 introduces a decision procedure for lightweight inclusions, based on the notion of constraint graphs, and discusses the problem of minimizing a set of lightweight inclusions. Section 5 shows how to compute the operations for lightweight ontologies. Section 6 describes the Protégé plug-in implementation. Section 7 summarizes related work. Section 8 contains the conclusions.

## 2 A Formal Framework

### 2.1 A Brief Review of Basic Concepts

The definition of the operations depends only on the notion of theory, which we introduce in the context of Description Logic (DL) (Baader and Nutt, 2003).

Briefly, a *vocabulary V* consists of a set of *atomic concepts*, a set of *atomic roles*, and the *bottom concept* $\bot$. A *language* in *V* is a set of strings, using symbols in *V*, whose definition depends on the specific variation of Description Logic adopted; the definition of the language typically includes definitions for the set of *concept descriptions in V* and for the set of *role descriptions in V*.

An *inclusion in V* is a statement of the form $u \sqsubseteq v$, where $u$ and $v$ both are concept descriptions in *V* or both are role descriptions in *V*. We use $u \equiv v$ (*equivalence*) as an abbreviation for the pair of inclusions $u \sqsubseteq v$ and $v \sqsubseteq u$.

An *interpretation* $\boldsymbol{s}$ for $V$ consists of a nonempty set $\Delta^s$, the *domain* of $\boldsymbol{s}$, whose elements are called *individuals*, and an *interpretation function*, also denoted $\boldsymbol{s}$, where:

$\boldsymbol{s}(\bot) = \varnothing$

$\boldsymbol{s}(A) \subseteq \Delta^s$ for each atomic concept $A$ in $V$

$\boldsymbol{s}(P) \subseteq \Delta^s \times \Delta^s$ for each atomic role $P$ in $V$

The function $\boldsymbol{s}$ is extended to role and concept descriptions in $V$. The exact definition again depends on the specific variation of Description Logic adopted. We use $\boldsymbol{s}(e)$ to indicate the value that $\boldsymbol{s}$ assigns to a concept description or a role description $e$ in $V$.

Let $\sigma$ and $\sigma'$ be two inclusions in $V$ and $\Sigma$ be a set of inclusions in $V$. Assume that $\sigma$ is of the form $u \sqsubseteq v$. We say that:

- $\boldsymbol{s}$ *satisfies* $\sigma$ or $\boldsymbol{s}$ is a *model* of $\sigma$, denoted $\boldsymbol{s} \vDash \sigma$, iff $\boldsymbol{s}(u) \subseteq \boldsymbol{s}(v)$.
- $\boldsymbol{s}$ *satisfies* $\Sigma$ or $\boldsymbol{s}$ is a *model* of $\Sigma$, denoted $\boldsymbol{s} \vDash \Sigma$, iff $\boldsymbol{s}$ satisfies all inclusions in $\Sigma$.
- $\sigma$ is *valid*, denoted $\vDash \sigma$, iff any interpretation for $V$ satisfies $\sigma$.
- $\sigma$ and $\sigma'$ are *tautologically equivalent* iff any model of $\sigma$ is a model of $\sigma'$ and vice-versa.
- $\Sigma$ *logically implies* $\sigma$, or $\sigma$ is a *logical consequence* of $\Sigma$, denoted $\Sigma \vDash \sigma$, iff any model of $\Sigma$ satisfies $\sigma$.
- $\Sigma$ is *satisfiable* or *consistent* iff there is a model of $\Sigma$.

The *theory* of $\Sigma$ in $V$, denoted $\tau[\Sigma]$, is the set of all inclusions in $V$ that are logical consequences of $\Sigma$. We say that two sets of inclusions, $\Gamma$ and $\Theta$, are *equivalent*, denoted $\Gamma \equiv \Theta$, iff $\tau[\Gamma]=\tau[\Theta]$.

Finally, an *ontology* is a pair $\boldsymbol{O}=(V,\Sigma)$ such that $V$ is a finite vocabulary, whose atomic concepts and atomic roles are called *classes* and *properties* of $\boldsymbol{O}$, respectively, and $\Sigma$ is a set of inclusions in $V$, called the *constraints* of $\boldsymbol{O}$. Two ontologies $O_1 = (V_1,\Sigma_1)$ and $O_2 = (V_2,\Sigma_2)$ are *equivalent*, denoted $O_1 \equiv O_2$, iff $\Sigma_1$ and $\Sigma_2$ are equivalent.

## 2.2 Definition of the Ontology Operations

In this section, we introduce a collection of operations over ontologies, whose definition is not restricted to any specific variation of DL.

**Definition 1**: Let $O_1 = (V_1,\Sigma_1)$ and $O_2 = (V_2,\Sigma_2)$ be two ontologies, $W$ be a subset of $V_1$, and $\Psi$ be a set of constraints in $V_1$.

(i) The *projection* of $O_1 = (V_1,\Sigma_1)$ over $W$, denoted $\pi[W](O_1)$, returns the ontology $O_P = (V_P,\Sigma_P)$, where $V_P = W$ and $\Sigma_P$ is the subset of the constraints in $\tau[\Sigma_1]$ that use only classes and properties in $W$.

(ii) The *deprecation* of $\Psi$ from $O_1 = (V_1,\Sigma_1)$, denoted $\delta[\Psi](O_1)$, returns the ontology $O_D = (V_D,\Sigma_D)$, where $V_D = V_1$ and $\Sigma_D = \Sigma_1 - \Psi$.

(iii) The *union* of $O_1 = (V_1,\Sigma_1)$ and $O_2 = (V_2,\Sigma_2)$, denoted $O_1 \cup O_2$, returns the ontology $O_U = (V_U,\Sigma_U)$, where $V_U = V_1 \cup V_2$ and $\Sigma_U = \Sigma_1 \cup \Sigma_2$.

(iv) The *intersection* of $O_1 = (V_1,\Sigma_1)$ and $O_2 = (V_2,\Sigma_2)$, denoted $O_1 \cap O_2$, returns the ontology $O_N = (V_N,\Sigma_N)$, where $V_N = V_1 \cap V_2$ and $\Sigma_N = \tau[\Sigma_1] \cap \tau[\Sigma_2]$.

(v) The *difference* of $O_1 = (V_1,\Sigma_1)$ and $O_2 = (V_2,\Sigma_2)$, denoted $O_1 - O_2$, returns the ontology $O_F = (V_F,\Sigma_F)$, where $V_F = V_1$ and $\Sigma_F = \tau[\Sigma_1] - \tau[\Sigma_2]$. □

Section 3 presents concrete examples of the operations. At this point, we observe that deprecation does not reduce to difference since, in general, we have

$$\tau[\Sigma_D] = \tau[\Sigma_1 - \Psi] \neq \tau[\Sigma_1] - \tau[\Psi]$$

Using the notation in Definition 1, we also observe that

$$\tau[\Sigma_U] = \tau[\Sigma_1 \cup \Sigma_2] \supseteq \tau[\Sigma_1] \cup \tau[\Psi]$$

$$\tau[\Sigma_N] = \tau[\tau[\Sigma_1] \cap \tau[\Sigma_2]] = \tau[\Sigma_1] \cap \tau[\Psi]$$

$$\tau[\Sigma_N] = \tau[\tau[\Sigma_1] - \tau[\Sigma_2]] \neq \tau[\Sigma_1] - \tau[\Psi]$$

In an earlier work (Casanova et al., 2011), we introduced the notions of open and closed fragments of an ontology. The operations capture these notions as follows. Let $O_1 = (V_1,\Sigma_1)$ be an ontology and $W$ be a subset of $V_1$. The *open fragment* of $O_1$ *defined by* $W$ is the projection $\pi[W](O_1)$ of $O_1$ over $W$, and the *closed fragment* of $O_1$ *defined by* $W$ is expressed as $\pi[W](O_1) \cup O_2$, where $O_2 = (V_1, \Phi)$ and $\Phi$ contains an inclusion of the form $A \sqsubseteq \bot$, for each atomic concept $A$ in $V_1$, but not in $W$, and an inclusion of the form $(\geq 1\ P) \sqsubseteq \bot$, for each atomic role $P$ in $V_1$, but not in $W$ (see Section 2.3 for the definition of $(\geq 1\ P)$); these inclusions force the interpretations of $A$ and $P$ to be the empty set.

We note that the ontology $O$ that results from an operation is unique, by definition. However, there might be several ontologies that are equivalent to $O$. For example, if $O_P = (V_P,\Sigma_P)$ is the projection of $O_1$ on $W$, there might be several sets of constraints that are equivalent to the set of constraints in the theory of $O_1$ that use only terms in $W$. This simple observation will be helpful in Section 5, which addresses how to implement the operations.

Finally, we observe that we may generalize union, intersection and difference by considering a renaming of one or both vocabularies of the ontologies involved and appropriately renaming the terms that occur in the constraints when comparing the theories. This extension is used in the first example of Section 3.2, but it will not be considered further in this paper.

### 2.3 Lightweight Description Logic

The procedures that implement the operations, introduced in Section 5, assume that the inclusions meet certain restrictions, imposed by a variation of Description Logic, called *Lightweight Description Logic*, or *Lightweight DL*.

Lightweight DL is characterized by the following definitions and restrictions on the sets of concept descriptions, role descriptions and inclusions.

**Definition 2**: Let $V$ be a vocabulary.

(i) A *lightweight role description* in $V$ is an atomic role $P$ in $V$ or a string of the form $P^-$ (*inverse role*), where $P$ is an atomic role in $V$.

(ii) A *lightweight basic concept description* in $V$ is the bottom concept $\bot$, an atomic concept in $V$, or an *at-least restriction* of the form $(\geq n\ p)$, where $p$ is a lightweight role description in $V$ and $n$ is a positive integer.

(iii) A *lightweight concept description* in $V$ is a lightweight basic concept description in $V$, or a *lightweight negated concept* of the form $\neg e$, where $e$ is a lightweight basic concept description in $V$.

(iv) A *lightweight inclusion* in $V$ is a string of one of the forms:

   (a) $e \sqsubseteq f$, where $e$ is an atomic concept or an at-least restriction in $V$ and $f$ is the bottom concept $\bot$, an atomic concept in $V$, or an at-least restriction.

   (b) $e \sqsubseteq \neg f$, where $e$ and $f$ are atomic concepts or at-least restrictions in $V$. □

**Definition 3**: Let *V* be a vocabulary and ***s*** be an interpretation for *V*. The function ***s*** is extended to lightweight role and concept descriptions in *V* as follows (where *P* is an atomic role, *e* is a lightweight basic concept description and *p* is a lightweight role description):

(i) $\boldsymbol{s}(P^{-}) = \boldsymbol{s}(P)^{-}$ (the inverse of $\boldsymbol{s}(P)$)

(ii) $\boldsymbol{s}(\neg e) = \Delta^{s} - \boldsymbol{s}(e)$ (the complement of $\boldsymbol{s}(e)$ with respect to $\Delta^{s}$)

(iii) $\boldsymbol{s}(\geq n\ p)=\{I \in \Delta^{s} \,/\, card(\{J \in \Delta^{s} \,/\, (I,J) \in \boldsymbol{s}(p)\}) \geq n\}$
(the set of individuals that $\boldsymbol{s}(p)$ relates to at least *n* distinct individuals, where *card*(*S*) denotes the cardinality of a set *S*). □

Since lightweight inclusions are a special case of inclusions, the notion of satisfiability, etc. remains as in Section 2.

We use the following abbreviations, where *p* is a lightweight role description:

- "⊤" (*universal concept*) for "¬⊥"
- "$\exists p$" (*existential quantification*) for "($\geq 1\ p$)"
- "($\leq n\ p$)" (*at-most restriction*) for "$\neg(\geq n{+}1\ p)$"

By an *unabbreviated* concept description we mean a concept description that does not use such abbreviations. Care must be taken to eliminate the abbreviated concept descriptions before checking if an inclusion is indeed a lightweight inclusion. Also, in view of the restrictions in Definition 2(iv), a *lightweight equivalence* $e \equiv f$ is such that *e* and *f* both are atomic concepts or at-least restrictions in *V*.

**Definition 4**: An ontology $\boldsymbol{O}=(V,\Sigma)$ is a *lightweight ontology* iff $\Sigma$ is a set of lightweight inclusions in *V*. □

We conclude this section with brief remarks on the expressiveness of lightweight inclusions.

Let *C* and *D* be atomic concepts, *p* and *q* be lightweight role descriptions and *m* and *n* be positive integers. According to Definition 2(iv), inclusions of the following forms are lightweight inclusions:

| | | | |
|---|---|---|---|
| (1) | $C \sqsubseteq \bot$ | $C \sqsubseteq D$ | $C \sqsubseteq (\geq m\ p)$ |
| | $(\geq n\ q) \sqsubseteq \bot$ | $(\geq n\ q) \sqsubseteq D$ | $(\geq n\ q) \sqsubseteq (\geq m\ p)$ |
| (2) | $C \sqsubseteq \neg D$ | $C \sqsubseteq \neg(\geq m\ p)$ | |
| | $(\geq n\ q) \sqsubseteq \neg D$ | $(\geq n\ q) \sqsubseteq \neg(\geq m\ p)$ | |

Lightweight inclusions are therefore sufficiently expressive to cover the simplest types of constraints used in conceptual modeling, as summarized in Table 1.

Let *e* and *f* be lightweight basic concept descriptions. Inclusions of the following forms are not lightweight inclusions:

| | | | |
|---|---|---|---|
| (3) | $\bot \sqsubseteq f$ | $\bot \sqsubseteq \neg f$ | $e \sqsubseteq \neg\bot$ |
| (4) | $\neg f \sqsubseteq \neg e$ | | |
| (5) | $\neg e \sqsubseteq f$ | | |

However, we note that

(6) $\bot \sqsubseteq f$, $\bot \sqsubseteq \neg f$, $e \sqsubseteq \neg\bot$ are valid (satisfiable by any interpretation)

(7) $\neg f \sqsubseteq \neg e$ is tautologically equivalent to $e \sqsubseteq f$

Therefore, when defining an ontology, inclusions as in (3) can be ignored, since they are vacuous constraints, and inclusions of the form $\neg f \sqsubseteq \neg e$ can be replaced by $e \sqsubseteq f$.

**Table 1.** Common constraint types used in conceptual modeling.

| Constraint Type | Abbreviated form | Unabbreviated form | Informal semantics |
|---|---|---|---|
| *Domain Constraint* | $\exists P \sqsubseteq C$ | $(\geq 1\ P) \sqsubseteq C$ | property $P$ has class $C$ as domain, that is, if $(a,b)$ is a pair in $P$, then $a$ is an individual in $C$ |
| *Range Constraint* | $\exists P^{-} \sqsubseteq C$ | $(\geq 1\ P^{-}) \sqsubseteq C$ | property $P$ has class $C$ as range, that is, if $(a,b)$ is a pair in $P$, then $b$ is an individual in $C$ |
| *minCardinality Constraint* | | $C \sqsubseteq (\geq k\ P)$ or $C \sqsubseteq (\geq k\ P^{-})$ | property $P$ or its inverse $P^{-}$ maps each individual in class $C$ to at least $k$ distinct individuals |
| *maxCardinality Constraint* | $C \sqsubseteq (\leq k\ P)$ or $C \sqsubseteq (\leq k\ P^{-})$ | $C \sqsubseteq \neg(\geq k{+}1\ P)$ or $C \sqsubseteq \neg(\geq k{+}1\ P^{-})$ | property $P$ or its inverse $P^{-}$ maps each individual in class $C$ to at most $k$ distinct individuals |
| *Subset Constraint* | | $C \sqsubseteq D$ | each individual in $C$ is also in $D$, that is, class $C$ denotes a subset of class $D$ |
| *Disjointness Constraint* | | $C \sqsubseteq \neg D$ | no individual is in both $C$ and $D$, that is, classes $C$ and $D$ are disjoint |

Finally, we remark that the definitions of $DL\text{-}Lite^{N}_{core}$ inclusions (Artale et al., 2009) and lightweight inclusions differ only in that the latter, but not the former, rules out inclusions of the forms in (3). However, this is semantically immaterial since, given a set $\Sigma$ of $DL\text{-}Lite^{N}_{core}$ inclusions, we can always drop from $\Sigma$ inclusions of the forms in (3) without affecting the theory of $\Sigma$, in view of (6). On the other hand, inclusion of the forms in (3) would unnecessarily complicate the structural proof procedure introduced in Section 4.1, based on Theorem 1.

## 3 Examples of the Operations

### 3.1 Projection, Deprecation and Union

Projection allows the designer to define a set $W$ containing just a few terms from the vocabulary of an ontology and retain the semantics of the terms in $W$ through the constraints, derivable from those of the ontology, that apply to the terms in $W$. Deprecation simply allows the designer to drop constraints from an ontology. Finally, union allows the designer to combine two ontologies. These three operations offer the designer powerful tools to (partially) reuse vocabularies and to preserve the semantics of the terms. In this section, we further motivate this argument with the help of an example that uses the *Music Ontology* (Raimond and Giasson, 2010).

The *Music Ontology* (MO) provides concepts and properties to describe artists, albums, tracks, performances, arrangements, etc. It is used by several Linked Data sources, including MusicBrainz and BBC Music. The Music Ontology RDF schema uses terms from the *Friend of a Friend* (FOAF) (Brickley and Miller, 2010) and the XML Schema (XSD) vocabularies. We respectively adopt the prefixes "mo:", "foaf:" and "xsd:" to refer to these vocabularies.

Figure 1 shows the class hierarchies of MO rooted at classes foaf:Agent and foaf:Person. Let us focus on this fragment of MO.

We first recall that FOAF has a constraint informally formulated as:

foaf:Person and foaf:Organization are disjoint classes

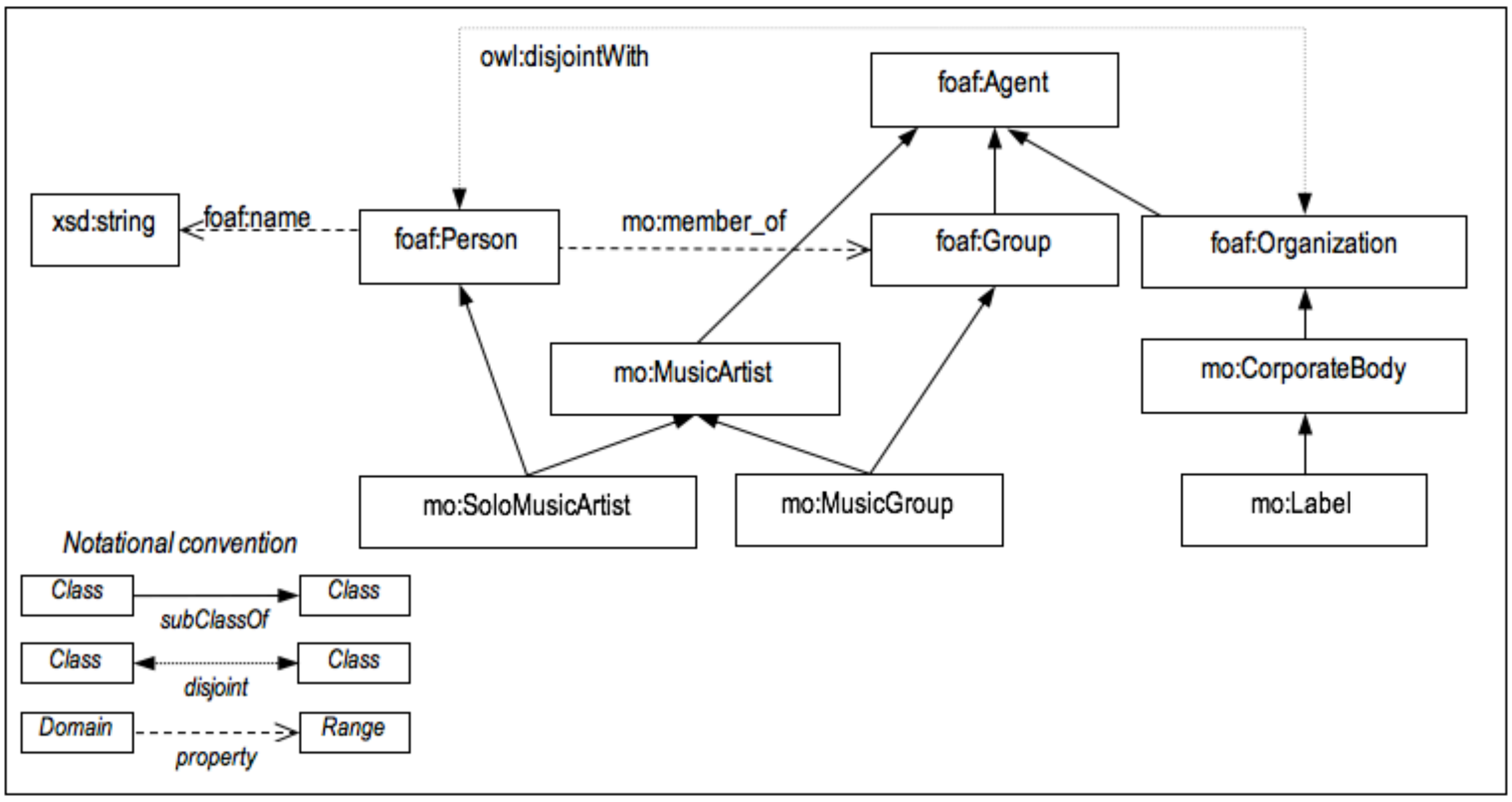

**Figure 1.** The class hierarchies of *MO* rooted at classes foaf:Agent and foaf:Person.

Let $V_1$ be the following set of terms from the FOAF and the XSD vocabularies, and let $V_2$ contain the rest of the terms that appear in Figure 1:

$V_1$ = {foaf:Agent, foaf:Person, foaf:Group, foaf:Organization, foaf:name, xsd:string}
$V_2$ = {mo:MusicArtist, mo:CorporateBody, mo:SoloMusicArtist, mo:MusicGroup, mo:Label, mo:member_of}

Let $\boldsymbol{O_1} = (V_1, \Sigma_1)$ be the ontology obtained by the *projection* of FOAF over $V_1$, denoted $\pi[V_1](FOAF)$ and defined in such a way that $\Sigma_1$ is the set of constraints over $V_1$ that are logical consequences of the constraints of FOAF:

$\Sigma_1$ = {(≥1 foaf:name) ⊑ foaf:Person, (≥1 foaf:name$^-$) ⊑ xsd:string,
foaf:Person ⊑ ¬foaf:Organization, foaf:Group ⊑ foaf:Agent,
foaf:Organization ⊑ foaf:Agent}

Let $\boldsymbol{O_2} = (V_2, \Sigma_2)$ be such that $\Sigma_2$ contains just the subset constraints over $V_2$ shown in Figure 1:

$\Sigma_2$ = {mo:SoloMusicArtist ⊑ mo:MusicArtist, mo:MusicGroup ⊑ mo:MusicArtist,
mo:Label ⊑ mo:CorporateBody}

Then, most of Figure 1 is captured by the *union* of $\boldsymbol{O_1}$ and $\boldsymbol{O_2}$, defined as the ontology $\boldsymbol{O_3} = (V_3, \Sigma_3)$, where $V_3 = V_1 \cup V_2$ and $\Sigma_3 = \Sigma_1 \cup \Sigma_2$.

The constraints shown in Figure 1, but not included in $\boldsymbol{O_3}$, are obtained by the union of $\boldsymbol{O_3} = (V_3, \Sigma_3)$ with $\boldsymbol{O_4} = (V_3, \Sigma_4)$ (the ontologies have the same vocabulary), where

$\Sigma_4$ = {mo:SoloMusicArtist ⊑ foaf:Person, mo:MusicGroup ⊑ foaf:Group,
mo:MusicArtist ⊑ foaf:Agent, mo:CorporateBody ⊑ foaf:Organization,
(≥1 mo:member_of) ⊑ foaf:Person, (≥1 mo:member_of$^-$) ⊑ foaf:Group}

The union returns the ontology $\boldsymbol{O_5} = (V_5, \Sigma_5)$, where $V_5 = V_3$ and $\Sigma_5 = \Sigma_3 \cup \Sigma_4$. Finally, we construct $\boldsymbol{O_0} = (V_0, \Sigma_0)$, the ontology that corresponds to Figure 1, as:

$\boldsymbol{O_0} = ((\pi[V_1](FOAF) \cup \boldsymbol{O_2}) \cup \boldsymbol{O_4})$

The reader is invited to reflect upon the definition of $\boldsymbol{O_0}$. We contend that the expression defined using the operations provides a reasonable explanation of how $\boldsymbol{O_0}$ is constructed from FOAF and additional terms and constraints.

### 3.2 Intersection and Difference

Intersection and difference help the designer compare the expressive power of two ontologies $O_1=(V_1,\Sigma_1)$ and $O_2=(V_2,\Sigma_2)$. If the designer wants to know what the ontologies have in common, he uses intersection. On the other hand, if he is interested in what holds in $O_1$, but not in $O_2$, he should use difference.

To illustrate the use of intersection, we analyze two data sources from the scientific research domain, DBLP and Lattes. DBLP stores Computer Science bibliographic references – over half a million references – and links to researchers' homepages. Lattes is a database, organized by CNPq – the Brazilian Research Agency, storing researchers' CVs and research group descriptions. Assume that the Lattes vocabulary suffers a renaming where Document is mapped to Publication.

To simplify the discussion, Table 2 shows just a few constraints from each data source. Column (a) shows the DBLP constraints, Column (b), the Lattes constraints, and Column (c) the constraints in the intersection. For example, Line 1 of the table indicates that Article ⊑ Publication is a constraint in both ontologies, after Document is renamed to Publication, and hence is in their intersection. Line 7(b) indicates that ConferencePaper ⊑ Publication is a constraint of the Lattes ontology, again after Document is renamed to Publication; whereas Lines 1(a) and 3(a) implies that ConferencePaper ⊑ Publication is in the theory of the DBLP ontology; hence this constraint is also in the intersection of the ontologies, as shown in Line 7(c).

To illustrate the use of difference, consider a scenario where a domain specialist adopted the version of the FOAF ontology released on January 1st, 2010 (call it FOAF1). However, on August 9th, 2010, a new release of the FOAF ontology was published (call it FOAF2). The specialist then wants to verify what changed from one version to the other. He can then take the difference between FOAF1 and FOAF2 (and vice-versa).

**Table 2**. Partial Intersection of the DBLP and Lattes ontologies.

| | (a) DBLP | (b) Lattes | (c) Intersection |
|---|---|---|---|
| 1 | Article ⊑ Publication | Article ⊑ Document | Article ⊑ Publication |
| 2 | Conference ⊑ Event | Book ⊑ Document | |
| 3 | ConferencePaper ⊑ Article | Collection ⊑ Document | |
| 4 | Continent ⊑ Place | Phdthesis ⊑ Document | |
| 5 | Proceedings ⊑ Publication | Proceedings ⊑ Document | Proceedings ⊑ Publication |
| 6 | Professor ⊑ Person | Series ⊑ Document | |
| 7 | | ConferencePaper ⊑ Document | ConferencePaper ⊑ Publication |

Table 3 shows the (partial) difference between FOAF1 and FOAF2. Line 2(c) indicates that the constraint Project ⊑ ¬Image is in the difference between FOAF1 and FOAF2. Indeed, since Image ⊑ Document is in the theory of FOAF1, we have that ¬Document ⊑ ¬Image is also in the theory of FOAF1. Hence, since Project ⊑ ¬Document is in the theory of FOAF1 (in fact, it is a constraint of FOAF1, according to Line 2(a)), we also have that Project ⊑ ¬Image is in the theory of FOAF1. However, this constraint is not in the theory of FOAF2. Likewise, Line 4(c) indicates that Organization ⊑ ¬Image is in the theory of FOAF1, but not in the theory of FOAF2. Finally, Line 6(a) indicates that FOAF1 has a constraint, Image ⊑ Document, which is not in the theory of FOAF2.

## 4 A Decision Procedure for Lightweight Inclusions

In this section, we review a decision procedure for lightweight inclusions, based on the notion of constraint graphs (Casanova et al., 2010). We also discuss the problem of minimizing a set of lightweight inclusions, which affects the implementation of the ontology operations. We stress that the concepts introduced in this section refer only to lightweight inclusions. Therefore, we often omit explicit reference to this variation of DL in what follows, a simplification that the reader must bear in mind.

**Table 3**. Partial difference between two versions of the FOAF ontology.

| | (a) FOAF1 (January 1st, 2010) | (b) FOAF2 (August 9th, 2010) | (c) Difference |
|---|---|---|---|
| 1 | | Agent ⊑ ¬Document | |
| 2 | Project ⊑ ¬Document | Project ⊑ ¬Document | Project ⊑ ¬Image |
| 3 | | Person ⊑ ¬Document | |
| 4 | Organization ⊑ ¬Document | Organization ⊑ ¬Document | Organization ⊑ ¬Image |
| 5 | Group ⊑ Agent | Group ⊑ Agent | |
| 6 | Image ⊑ Document | | Image ⊑ Document |

### 4.1 Constraint Graphs

We say that the *complement* of a basic concept description $b$ is $\neg b$, and vice-versa. If $e$ is a basic concept description, or the negation of a basic concept description, then $\bar{e}$ denotes the complement of $e$.

Let $\Sigma$ be a set of lightweight inclusions and $\Omega$ be a set of lightweight concept descriptions.

**Definition 5**: The labeled graph $g(\Sigma,\Omega)=(\gamma,\delta,\kappa)$ that *captures* $\Sigma$ and $\Omega$, where $\kappa$ labels each node with a concept description, is defined as follows:

(i) For each concept description $e$ that occurs on the right- or left-hand side of an inclusion in $\Sigma$, or that occurs in $\Omega$, there is exactly one node in $\gamma$ labeled with $e$. If necessary, the set of nodes is augmented with new nodes so that:
  (a) For each atomic concept $C$ that occurs in $\Sigma$ or in $\Omega$, there is exactly one node in $\gamma$ labeled with $C$.
  (b) For each atomic role $P$ that occurs in $\Sigma$ or in $\Omega$, there is exactly one node in $\gamma$ labeled with ($\geq 1$ $P$) and exactly one node labeled with ($\geq 1$ $P^-$).

(ii) If there is a node in $\gamma$ labeled with a concept description $e$, then there must be exactly one node in $\gamma$ labeled with $\bar{e}$.

(iii) For each inclusion $e \sqsubseteq f$ in $\Sigma$, there is an arc $(M,N)$ in $\delta$, where $M$ and $N$ are the nodes labeled with $e$ and $f$, respectively.

(iv) If there are nodes $M$ and $N$ in $\gamma$ labeled with ($\geq m$ $p$) and ($\geq n$ $p$) such that $m<n$, where $p$ is either $P$ or $P^-$, then there is an arc $(N,M)$ in $\delta$. Such arc are called *tautological arcs*.

(v) If there is an arc $(M,N)$ in $\delta$ such that $M$ and $N$ are labeled with $e$ and $f$, respectively, then there is an arc $(K,L)$ in $\delta$ such that $K$ and $L$ are the nodes labeled with $\bar{f}$ and $\bar{e}$, respectively.

(vi) These are the only nodes and arcs of $g(\Sigma,\Omega)$.

When $\Omega$ is the empty set, we simply write $g(\Sigma)$ and say that $g(\Sigma)$ is the graph that *captures* $\Sigma$. □

**Definition 6**: The *constraint graph* for $\Sigma$ and $\Omega$ is the labeled graph $G(\Sigma,\Omega)=(\eta,\varepsilon,\lambda)$, where $\lambda$ labels each node with a set of concept descriptions. The graph $G(\Sigma,\Omega)$ is defined by collapsing each strongly connected component of $g(\Sigma,\Omega)$ into a single node, labeled with the set of concept descriptions that previously labeled the nodes in the strongly connected component. When $\Omega$ is the empty set, we simply write $G(\Sigma)$ and say that $G(\Sigma)$ is the *constraint graph* for $\Sigma$. □

If a node $K$ of $G(\Sigma,\Omega)$ is labeled with $e$, then $\overline{K}$ denotes the node labeled with $\bar{e}$; we say that $K$ and $\overline{K}$ are *dual nodes* and $(M,N)$ and $(\overline{N},\overline{M})$ are *dual arcs*.

**Example 1**: The fragment of the Music Ontology shown in Figure 1 is formalized as the *Agent-Person* ontology, ***APO*** = $(V_{APO},\Sigma_{APO})$, where

$V_{APO}$ = { foaf:Agent, foaf:Person, foaf:Group, foaf:Organization, mo:MusicArtist, mo:CorporateBody, mo:SoloMusicArtist, mo:MusicGroup, mo:Label, mo:member_of, foaf:name, xsd:string }

$\Sigma_{APO}$ = (the set of constraints is shown in Table 4)

**Table 4.** The constraints of the ontology ***APO*** (unabbreviated form).

| Constraint | Informal specification |
|---|---|
| (≥1 foaf:name) ⊑ foaf:Person<br>(≥1 foaf:name¯ ) ⊑ xsd:string<br>(≥1 mo:member_of) ⊑ foaf:Person<br>(≥1 mo:member_of¯ ) ⊑ foaf:Group | The domain of foaf:name is foaf:Person<br>The range of foaf:name is xsd:string<br>The domain of mo:member_of is foaf:Person<br>The range of mo:member_of is foaf:Group |
| mo:MusicArtist ⊑ foaf:Agent<br>foaf:Group ⊑ foaf:Agent<br>foaf:Organization ⊑ foaf:Agent<br>mo:SoloMusicArtist ⊑ foaf:Person<br>mo:SoloMusicArtist ⊑ mo:MusicArtist<br>mo:MusicGroup ⊑ mo:MusicArtist<br>mo:MusicGroup ⊑ foaf:Group<br>mo:CorporateBody ⊑ foaf:Organization<br>mo:Label ⊑ mo:CorporateBody | mo:MusicArtist is a subset of foaf:Agent<br>foaf:Group is a subset of foaf:Agent<br>foaf:Organization is a subset of foaf:Agent<br>mo:SoloMusicArtist is a subset of foaf:Person<br>mo:SoloMusicArtist is a subset of mo:MusicArtist<br>mo:MusicGroup is a subset of mo:MusicArtist<br>mo:MusicGroup is a subset of foaf:Group<br>mo:CorporateBody is a subset of foaf:Organization<br>mo:Label is a subset of mo:CorporateBody |
| foaf:Person ⊑ ¬foaf:Organization | foaf:Person and foaf:Organization are disjoint |

Figure 2 depicts the constraint graph $g(\Sigma_{APO})$ for $\Sigma_{APO}$. Since $g(\Sigma_{APO})$ has no strongly connected components, $g(\Sigma_{APO})$ and $G(\Sigma_{APO})$ are in fact the same graph. Note that there is a path from the node labeled with mo:Label to the node labeled with ¬(≥1 mo:member_of), which indicates that mo:Label ⊑ ¬(≥1 mo:member_of) is a logical consequence of $\Sigma_{APO}$. This logical implication would not be captured if we constructed the graph with just the concept descriptions that occur in $\Sigma_{APO}$. Hence, it provides an example of why we need Conditions (ii) and (v) in Definition 5.

We use $K\rightarrow M$ to indicate that there is a path in $G(\Sigma,\Omega)$ from $K$ to $M$. Also, as a convenience, a *path of length 0* is a path consisting of a single node.

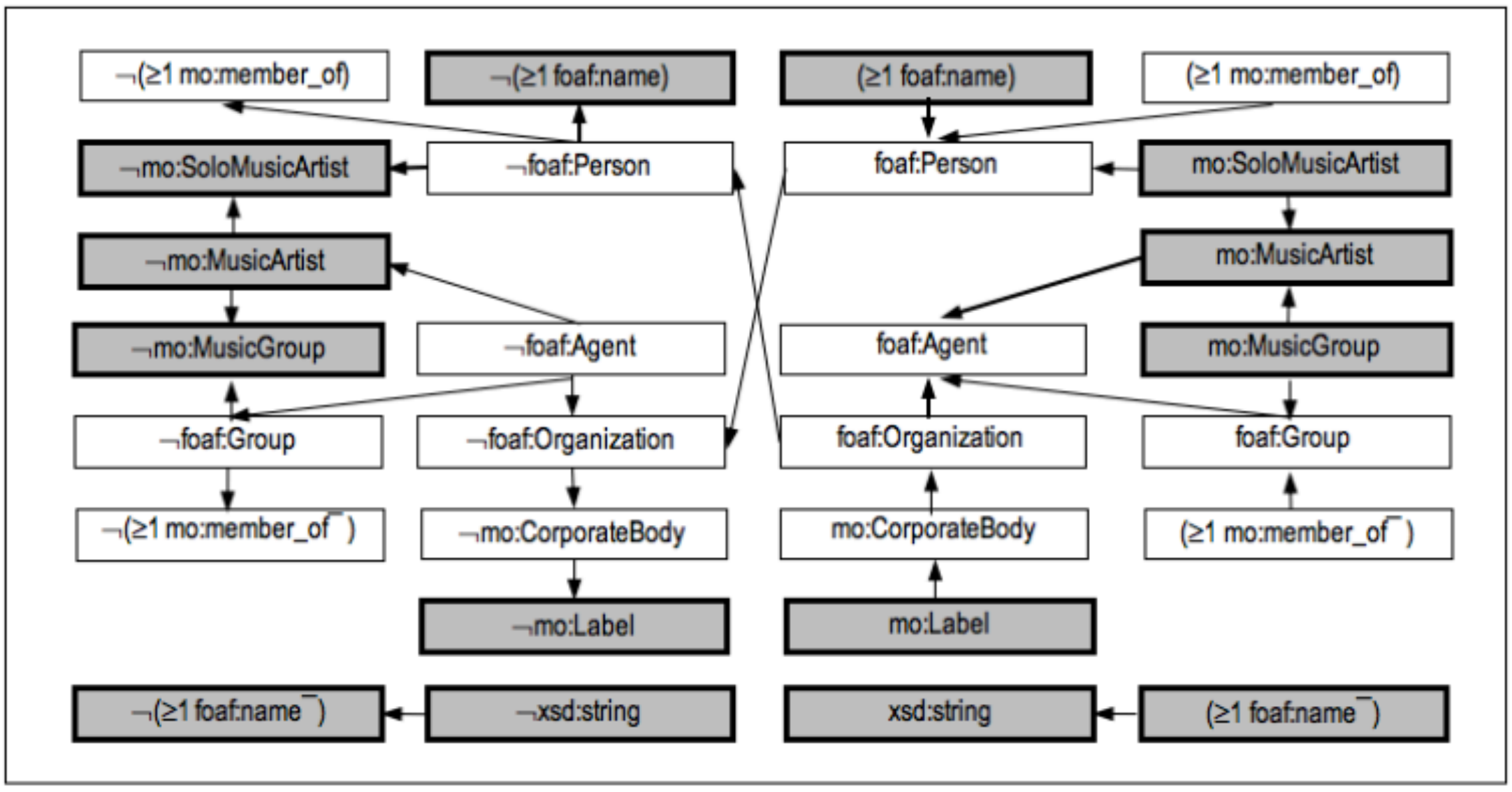

**Figure 2.** The constraint graph $G(\Sigma_{APO})$ for $\Sigma_{APO}$.

**Definition 7**: Let $G(\Sigma,\Omega)=(\eta,\varepsilon,\lambda)$ be the constraint graph for $\Sigma$ and $\Omega$.

(i) We say that a node $K$ of $G(\Sigma,\Omega)$ is a *⊥-node of rank 0* iff

(a) $K$ is labeled with ⊥, or

(b) $K$ is not labeled with ⊥, there is no ⊥-node $L$ of rank 0 such that $(K,L)$ is an arc of $G(\Sigma,\Omega)$, and there are nodes $M$ and $N$, not necessarily distinct from $K$, and a basic concept description $b$ such that $M$ and $N$ are labeled with $b$ and $\neg b$, respectively, and $K \rightarrow M$ and $K \rightarrow N$.

(ii) For each positive integer $n$, we say that a node $K$ of $G(\Sigma,\Omega)$ is a *⊥-node of rank n* iff $K$ is not a ⊥-node of rank $m$, with $m<n$, and there is a ⊥-node $L$ of rank $n$-$1$ such that

(a) $(K,L)$ is an arc of $G(\Sigma,\Omega)$, or

(b) $L$ is labeled with (≥1 $P^{-}$) and $K$ is labeled with (≥1 $P$), or

(c) $L$ is labeled with (≥1 $P$) and $K$ is labeled with (≥1 $P^{-}$). □

Case (ii-b) captures the fact that, given an interpretation $\mathbf{s}$, if $\mathbf{s}((\geq 1\ P^{-}))=\varnothing$, then $\mathbf{s}(P)=\mathbf{s}((\geq 1\ P))=\varnothing$. Case (ii-c) follows likewise, when $\mathbf{s}((\geq 1\ P))=\varnothing$. In view of these cases, the notion of rank is necessary to avoid a circular definition.

**Definition 8**: Let $G(\Sigma,\Omega)=(\eta,\varepsilon,\lambda)$ be the constraint graph for $\Sigma$ and $\Omega$. Let $K$ be a node of $G(\Sigma,\Omega)$. We say that $K$ is a *⊥-node* iff $K$ is a ⊥-node with rank $n$, for some non-negative integer $n$. We also say that $K$ is a *⊤-node* iff $\overline{K}$ is a ⊥-node. □

To simplify the procedures in Sections 4.2 and 5, we label the ⊥-nodes and the ⊤-nodes of a constraint graph with "⊥-node" and "⊤-node", respectively. We refer to these new labels as *tags*, to avoid confusion with the labels introduced in Definition 5.

**Definition 9**: The *tagged* constraint graph for $\Sigma$ and $\Omega$ is the constraint graph for $\Sigma$ and $\Omega$, with the ⊥-nodes and the ⊤-nodes tagged with "⊥-node" and "⊤-node", respectively. □

In what follows, let $G(\Sigma,\Omega)$ be the constraint graph for a set $\Sigma$ of lightweight inclusions and a set $\Omega$ of lightweight concept descriptions. Propositions 1, 2 and 3 capture some simple properties of $G(\Sigma,\Omega)$. We only provide brief proof sketches for these propositions and refer the reader to (Casanova et al., 2010) for the details.

**Proposition 1** (Duality properties):

(i) For any pair of nodes $M$ and $N$ of $G(\Sigma,\Omega)$, we have that $M \rightarrow N$ iff $\overline{N} \rightarrow \overline{M}$.

(ii) For any node $M$ of $G(\Sigma,\Omega)$, for any concept description $e$, we have that $e$ labels $M$ iff $\bar{e}$ labels $\overline{M}$.

(iii) For any node $M$ of $G(\Sigma,\Omega)$, we have that $M$ is a $\bot$-node iff $\overline{M}$ is a $\top$-node.

**Proof sketch**

Property (i) follows from Definition 5(v) and Definition 6. Property (ii) follows from Definition 5(ii) and Definition 6. Finally, Property (iii) follows from Definition 8. □

**Proposition 2** (Basic properties of constraint graphs):

(i) $G(\Sigma,\Omega)$ is acyclic.

(ii) For any lightweight concept description $e$ that occurs in $\Sigma$ or $\Omega$, there is just one node $M$ of $G(\Sigma,\Omega)$ such that $e$ labels $M$.

(iii) For any node $M$ of $G(\Sigma,\Omega)$,

   (a) $M$ is labeled only with $\bot$, or
   (b) $M$ is labeled only with $\top$, or
   (c) $M$ is labeled only with atomic concepts and at-least restrictions, or
   (d) $M$ is labeled only with negated atomic concepts and negated at-least restrictions.

(iv) For any node $M$ of $G(\Sigma,\Omega)$, if $M$ is a $\bot$-node and $M$ is not labeled with $\bot$, then $M$ is labeled only with atomic concepts and at-least restrictions.

(v) For any node $M$ of $G(\Sigma,\Omega)$, if $M$ is a $\top$-node and $M$ is not labeled with $\top$, then $K$ is labeled only with negated atomic concepts and negated at-least restrictions.

(vi) For any arc $(M,N)$ of $G(\Sigma,\Omega)$,

   (a) $M$ and $N$ are both labeled only with atomic concepts and at-least restrictions, or
   (b) $M$ is labeled only with atomic concepts and at-least restrictions and $N$ is labeled only with $\bot$, or
   (c) $M$ is labeled only with atomic concepts and at-least restrictions and $N$ is labeled only with negated atomic concepts and negated at-least restrictions, or
   (d) $M$ and $N$ are both labeled only with negated atomic concepts and negated at-least restrictions, or
   (e) $M$ is labeled only with $\top$ and $N$ is labeled only with negated atomic concepts and negated at-least restrictions.

**Proof sketch**

Property (i) follows since, by Definition 6, $G(\Sigma,\Omega)$ is defined by collapsing each strongly connected component of $g(\Sigma,\Omega)$ into a single node.

Property (ii) follows from Definition 5(i)(ii) and Definition 6.

To prove Properties (iii) to (vi), we first observe that, by Definition 2(iv):

(*) $\neg e \sqsubseteq f$ is not a lightweight inclusion, where $e$ and $f$ are lightweight basic concept descriptions

Then, from (*) and Definition 5, all nodes in a strongly connected component of $g(\Sigma,\Omega)$ are labeled only with atomic concepts and at-least restrictions or only with negated atomic concepts and negated at-least restrictions. Hence, from Definition 6, we have that $G(\Sigma,\Omega)$ satisfies (iii).

Properties (iv) and (v) follow from Property (iii) and Definitions 7 and 8.

Property (vi) follows again from (*) and Definitions 5 and 6. □

Recall that the complement of a basic concept description $b$ is $\neg b$, and vice-versa. Also recall that, if $e$ is a basic concept description, or the negation of a basic concept description, then $\bar{e}$ denotes the complement of $e$.

**Proposition 3** (Consistency):

(i) For any pair of nodes $M$ and $N$ of $G(\Sigma,\Omega)$ such that $M$ is labeled only with atomic concepts and at-least restrictions, for any label $e$ of $M$, for any label $f$ of $N$, if $M \rightarrow N$ then $e \sqsubseteq f$ is a lightweight inclusion and $\Sigma \vDash e \sqsubseteq f$.

(ii) For any node $M$ of $G(\Sigma,\Omega)$ such that $M$ is labeled only with atomic concepts and at-least restrictions, for any label $e$ of $M$, if $M$ is a $\bot$-node, then $e \sqsubseteq \bot$ is a lightweight inclusion and $\Sigma \vDash e \sqsubseteq \bot$.

(iii) For any node $M$ of $G(\Sigma,\Omega)$ such that $M$ is labeled only with atomic concepts and at-least restrictions, for any pair $e$ and $f$ of labels of $M$, $e \equiv f$ is a lightweight equivalence and $\Sigma \vDash e \equiv f$.

**Proof sketch**

Property (i) follows by the transitivity of inclusions, Proposition 2 and Definitions 5 and 6.

Property (ii) follows from Proposition 2(iv), Property (i) and Definitions 7 and 8, if we observe that, if $\Sigma \vDash e \sqsubseteq b$ and $\Sigma \vDash e \sqsubseteq \neg b$ then $\Sigma \vDash e \sqsubseteq \bot$.

Property (iii) follows from Proposition 1 and Property (ii).

Property (iv) follows by the transitivity of inclusions, Proposition 2 and Definitions 5 and 6. □

Theorem 1 shows how to test logical implication for lightweight inclusions. In the "if" direction, Theorem 1 is just a restatement of Proposition 3. In the "only if" direction, the proof is far more complex that those of the previous propositions and can be found in (Casanova et al., 2010). Just as a reminder, a path from a node $M$ to a node $N$ has length 0 iff $M=N$.

**Theorem 1** (Completeness): Let $\Sigma$ be a set of lightweight inclusions and $e \sqsubseteq f$ be a lightweight inclusion. Let $\Omega=\{e,f\}$. Then, $\Sigma \vDash e \sqsubseteq f$ iff one of the following conditions holds:

(i) The node of $G(\Sigma,\Omega)$ labeled with $e$ is a $\bot$-node, or

(ii) The node of $G(\Sigma,\Omega)$ labeled with $f$ is a $\top$-node, or

(iii) There is a path in $G(\Sigma,\Omega)$, possibly with length 0, from the node labeled with $e$ to the node labeled with $f$. □

Theorem 1 leads to a simple decision procedure, **Implies**, to test if a lightweight inclusion $e \sqsubseteq f$ is a logical consequence of a set $\Sigma$ of lightweight inclusions (see Figure 3).

**Example 2**: Consider again the fragment of the Music Ontology formalized as the *Agent-Person* ontology, ***APO*** = $(V_{APO}, \Sigma_{APO})$, in Example 1. From Theorem 1, we have that the following inclusions are logical consequences of $\Sigma_{APO}$ (note that such inclusions are not in $\Sigma_{APO}$):

mo:Label ⊑ foaf:Organization	mo:Label ⊑ foaf:Agent
mo:Label ⊑ ¬foaf:Person	mo:Label ⊑ ¬mo:SoloMusicArtist
mo:Label ⊑ ¬(≥1 foaf:name)	mo:Label ⊑ ¬(≥1 mo:member_of) . □

### 4.2 Minimizing the Set of Constraints of a Lightweight Ontology

This section in devoted of the question of minimizing the set of constraints of a lightweight ontology. We argue that this question is quite similar to finding a *minimal equivalent graph (MEG)* of a graph *G*, defined as a graph *H* with a minimal set of edges such that the transitive closure of *G* and *H* are equal. This problem has a polynomial solution when *G* is acyclic and is NP-hard for strongly connected graphs (Aho, Garey and Ullman, 1972; Hsu, 1975; Khuller, Raghavachari and Young, 1975). Figure 3 contains all procedures developed to address this question, as well as procedures to construct constraint graphs and to test if a lightweight inclusion is a logical consequence of a set of lightweight inclusions, based on Theorem 1.

The **MinimizeGraph** procedure is based on the strategy of finding the MEG of a constraint graph. Step (2) of **MinimizeGraph** can be implemented in polynomial time (Aho, Garey and Ullman, 1972; Hsu, 1975), since *H* is acyclic (because so is *G*, by Proposition 2(i)). In view of Propositions 1 and 2, Step (2) considers just the nodes of *H* labeled only with atomic concepts and at-least restrictions. Furthermore, since Step (2) drops the dual arcs, *H* satisfies the properties listed in Propositions 1 and 2.

The **GenerateConstraints** procedure transforms graph *H* into a set of constraints $\Sigma_2$. Again, in view of Propositions 1 and 2, Step (3) of **GenerateConstraints** considers just the nodes of *H* labeled only with atomic concepts and at-least restrictions.

Finally, the **MinimizeConstraints** uses the previous procedure to transform a set of lightweight constraints $\Sigma_1$ and output an equivalent, minimal set of constraints $\Sigma_2$. The correctness of **MinimizeConstraints** is stated in Theorem 2.

**Theorem 2:** Let $\Sigma_1$ be a set of lightweight constraints and $\Sigma_2$ be the result of applying **MinimizeConstraints** to $\Sigma_1$. Then, $\Sigma_1$ and $\Sigma_2$ are equivalent, that is, $\tau[\Sigma_1]=\tau[\Sigma_2]$.

(See the appendix for a proof).

**ConstructGraph:**

**Input:** a set $\Sigma$ of lightweight inclusions and an optional lightweight inclusion $e \sqsubseteq f$
**Output:** the tagged constraint graph $G(\Sigma,\Omega)$

1. Construct the constraint graph $G(\Sigma,\Omega)$ for $\Sigma$ and $\Omega=\{e,f\}$, using Definition 6.
2. Tag $G(\Sigma,\Omega)$, using Definition 9.
3. Return $G(\Sigma,\Omega)$.

**Implies:**

**Input:** a lightweight inclusion $e \sqsubseteq f$ and a set $\Sigma$ of lightweight inclusions
**Output:** "True", if $e \sqsubseteq f$ is a logical consequence of a set $\Sigma$
"False", otherwise

1. Call **ConstructGraph** to construct the constraint graph $G(\Sigma,\Omega)$ for $\Sigma$ and $\Omega=\{e,f\}$.
2. Return "True" if
3. The node of $G(\Sigma,\Omega)$ labeled with $e$ is a $\bot$-node, or
4. The node of $G(\Sigma,\Omega)$ labeled with $f$ is a $\top$-node, or
5. There is a path in $G(\Sigma,\Omega)$, possibly with length 0, from the node labeled with $e$ to the node labeled with $f$.
6. Return "False", otherwise.

**MinimizeGraph:**

**Input:** a tagged constraint graph $G$
**Output:** a MEG $H$ of $G$

1. Initialize $H$ with the same nodes, arcs, labels and tags as $G$.
2. For each node $L$ of $H$ labeled only with atomic concepts and at-least restrictions,
3. For each arc $(L,M)$ in $H$,
4. For each node $N$ in $H$, do:
5. If there are arcs $(M,N)$ and $(L,N)$ in $H$
6. such that $(L,N)$ is not a tautological arc,
7. Drop from $H$ both the arc $(L,N)$ and the arc $(\overline{N},\overline{L})$ connecting the dual nodes of $L$ and $M$.

**GenerateConstraints:**

**Input:** a tagged constraint graph $H$
**Output:** a set of constraints $\Sigma_2$

1. Initialize $\Sigma_2$ to be the empty set.
2. Mark all arcs of $H$ as unprocessed.
3. For each node $M$ of $H$ labeled only with atomic concepts and at-least restrictions, do:
4. If $M$ is tagged as a "$\bot$-node", then
5. For each label $e$ of $M$,
6. Add to $\Sigma_2$ a constraint of the form $e \sqsubseteq \bot$.
7. If $M$ is not tagged as "$\bot$-node", then
8. Order the labels of $M$, creating a list $e_1,\ldots,e_n$, and
9. Add to $\Sigma_2$ the constraints $e_1 \sqsubseteq e_2$, $e_2 \sqsubseteq e_3$,…, $e_{n-1} \sqsubseteq e_n$ and $e_n \sqsubseteq e_1$.
10. For each arc $(M,N)$ of $H$ such that $(M,N)$ is unprocessed, do:
11. Select a label $e$ of $M$ and a label $f$ of $N$ and
12. Add to $\Sigma_2$ a constraint of the form $e \sqsubseteq f$.
13. Mark both $(M,N)$ and $(\overline{N},\overline{M})$ as processed.
14. Return $\Sigma_2$.

**MinimizeConstraints:**

**Input:** a set of lightweight constraints $\Sigma_1$
**Output:** an equivalent, minimal set of constraints $\Sigma_2$

1. Call **ConstructGraph** to construct the tagged constraint graph $G$ for $\Sigma_1$.
2. Call **MinimizeGraph** with $G$ to generate $H$.
3. Call **GenerateContraints** with $H$ to generate $\Sigma_2$.
4. Return $\Sigma_2$.

**Figure 3.** Basic procedures.

The following example illustrates how **MinimizeConstraints** operates.

**Example 3**: The *Person Music Group* ontology, ***PMG*** = $(V_{PMG}, \Sigma_{PMG})$, is such that:

$V_{PMG}$ = { foaf:Agent, foaf:Person, foaf:Group, mo:MusicGroup, mo:member_of }

$\Sigma_{PMG}$ = (the constraints in Table 5, with redundancies for the sake of our example)

**Table 5.** Constraints of the ***PMG*** ontology.

| Constraint | Informal specification |
|---|---|
| (≥1 mo:member_of) ⊑ foaf:Person<br>(≥1 mo:member_of¯) ⊑ foaf:Group<br>(≥1 mo:member_of¯) ⊑ foaf:Agent | The domain of mo:member_of is foaf:Person<br>The range of mo:member_of is foaf:Group<br>(Redundant constraint) |
| mo:MusicGroup ⊑ foaf:Group<br>mo:MusicGroup ⊑ foaf:Agent<br>foaf:Group ⊑ foaf:Agent | mo:MusicGroup is a subset of foaf:Group<br>(Redundant constraint)<br>foaf:Group is a subset of foaf:Agent |
| foaf:Person ⊑ ¬foaf:Agent | foaf:Person is disjoint from foaf:Agent |

Figure 4 shows the constraint graph $G(\Sigma_{PMG})$. **MinimizeConstraints** first creates a MEG *H* of $G(\Sigma_{PMG})$ by discarding the arcs indicated in Figure 4 as dashed lines. Next, **MinimizeConstraints** calls **GenerateConstraints**, which outputs the final set of constraints $\Theta_{PMG}$ shown in Table 6. Note that **GenerateConstraints** does not include, in the final set, the constraint

mo:MusicGroup ⊑ foaf:Agent

or the constraint

(≥1 mo:member_of¯) ⊑ foaf:Agent

and includes, in the final set, one of the constraints

foaf:Person ⊑ ¬foaf:Agent
foaf:Agent ⊑ ¬foaf:Person

but not both, because Step (3-b-ii-b) marks as processed both an arc and its dual in each iteration.

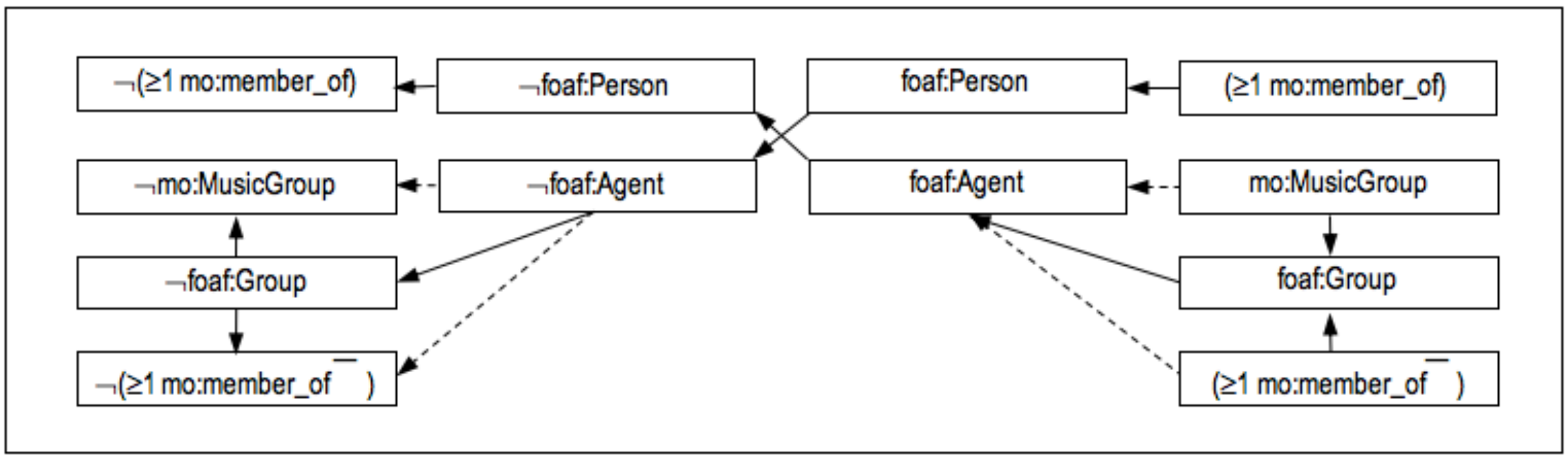

**Figure 4.** The constraint graph $G(\Sigma_{PMG})$ for $\Sigma_{PMG}$.

**Table 6**. Minimized set of Constraints $\Theta_{PMG}$ of the ***PMG*** ontology.

| Constraint | Informal specification |
|---|---|
| (≥1 mo:member_of) ⊑ foaf:Person<br>(≥1 mo:member_of¯) ⊑ foaf:Group | The domain of mo:member_of is foaf:Person<br>The range of mo:member_of is foaf:Group |
| mo:MusicGroup ⊑ foaf:Group<br>foaf:Group ⊑ foaf:Agent | mo:MusicGroup is a subset of foaf:Group<br>foaf:Group is a subset of foaf:Agent |
| foaf:Person ⊑ ¬foaf:Agent | foaf:Person is disjoint from foaf:Agent |

## 5 Implementation of the Operations

### 5.1 Overview

We start with a few observations that, albeit simple, impact the implementation of the operations. First, Definition 1 guarantees that each operation is a function, as expected, that is, each operation returns a unique result $O_R$ for each input. However, we consider acceptable that the implementation of an operation actually computes an ontology $O_E$ which is equivalent to $O_R$. Furthermore, if the input ontologies have a finite set of constraints, we require that the implementation returns an ontology that has a finite set of constraints. This may be problematic for projection, intersection and difference, whose definitions use the theories of the sets of constraints involved, rather than the sets of constraints themselves, as in the definition of deprecation and union.

Let $O_1 = (V_1,\Sigma_1)$ and $O_2 = (V_2,\Sigma_2)$ be two lightweight ontologies, $W$ be a subset of $V_1$ and $\Psi$ be a set of constraints in $V_1$. From the perspective of the difficulty of implementation, we may divide the operations into three groups:

*Group 1:* deprecation and union.
These operations have direct implementations from Definitions 1(ii) and (iii). Given $O_1$ and $\Psi$, the **Deprecation** procedure returns the ontology $O_D = (V_D,\Sigma_D)$, where $V_D = V_1$ and $\Sigma_D$ is the result of minimizing $\Sigma_1 - \Psi$. Given $O_1$ and $O_2$, the **Union** procedure returns the ontology $O_U = (V_U,\Sigma_U)$, where $V_U = V_1 \cup V_2$ and $\Sigma_U$ is the result of minimizing $\Sigma_1 \cup \Sigma_2$. Hence, these procedures are quite simple and will not be further discussed.

*Group 2:* projection and intersection.
These operations have implementations that depend on Theorem 1. The **Projection** procedure computes the projection of $O_1$ onto $W$ and follows directly from Definition 1(i), Theorem 1 and constraint minimization. The **Intersection** procedure likewise follows from Definition 1(iv). These procedures will be discussed in detail in Sections 5.2 and 5.3.

*Group 3:* difference.
This operation raises difficulties as discussed in Section 5.4.

### 5.2 Implementation of Projection

Let $O_1 = (V_1,\Sigma_1)$ be a lightweight ontology and $W$ be a subset of $V_1$. Recall that the projection of $O_1$ over $W$ is the ontology $O_P = (V_P, \Sigma_P)$, where $V_P = W$ and $\Sigma_P$ is the set of constraints in $\tau[\Sigma_1]$ that use only classes and properties in $W$.

Procedure **Projection** computes $\Gamma_P$ so that $\tau[\Gamma_P] = \tau[\Sigma_P]$. That is, given any lightweight inclusion $e \sqsubseteq f$ that involves only classes and properties in $W$, $e \sqsubseteq f$ is a logical consequence of $\Gamma_P$ iff $e \sqsubseteq f$ is a logical consequence of $\Sigma_1$. Note that this does not mean that $e \sqsubseteq f$ is a logical consequence of the subset of $\Sigma_1$ whose inclusions involve only classes and properties in $W$.

Procedure **Projection** works as follows:

(1) Construct $G(\Sigma_1)$, the tagged constraint graph for $\Sigma_1$.

(2) Construct $G^*(\Sigma_1)$, the transitive closure of $G(\Sigma_1)$. The nodes of $G^*(\Sigma_1)$ retain all labels and tags as in $G(\Sigma_1)$.

(3) Use $G^*(\Sigma_I)$ to create a graph $G_W$ by discarding all concept descriptions that label nodes of $G^*(\Sigma_I)$ and that involve classes and properties which are not in $W$; nodes that end up with no labels are discarded, as well as their adjacent arcs. The nodes of $G_W$ retain all tags as in $G^*(\Sigma_I)$.

(4) Call **MinimizeGraph** with $G_W$ to generate $H$.

(5) Call **GenerateContraints** with $H$ to generate $\Gamma_P$.

(6) Return $O_P = (W, \Gamma_P)$.

The correctness of **Projection** is established in Theorem 3, whose proof follows directly from Theorem 1. In particular, the transitive closure $G^*(\Sigma_I)$, generated in Step (2), is simply a convenient way to capture all paths in $G(\Sigma_I)$ required to apply Condition (iii) of Theorem 1.

**Theorem 3** (Correctness of **Projection**): Let $O_I = (V_I,\Sigma_I)$ be a lightweight ontology and $W$ be a subset of $V_I$. Let $O_P = (W, \Gamma_P)$ be the ontology that **Projection** returns for $O_I$ and $W$. Then, for any lightweight inclusion $e \sqsubseteq f$ that involves only classes and properties in $W$, we have that $\Sigma_I \models e \sqsubseteq f$ iff $\Gamma_P \models e \sqsubseteq f$. □

The following example illustrates how **Projection** operates.

**Example 2**: Consider the following vocabulary:

$V_{MAC}$ = {mo:MusicArtist, mo:SoloMusicArtist, mo:MusicGroup, mo:Label, foaf:name, xsd:string}

The *Music Artist Contract* ontology, $\boldsymbol{MAC} = (V_{MAC},\Sigma_{MAC})$, is defined as the projection of $\boldsymbol{APO} = (V_{APO},\Sigma_{APO})$, defined in Example 1, over $V_{MAC}$. The nodes labeled with expressions that use only classes and properties in $V_{MAC}$ are depicted in gray boxes with thicker frames in Figure 2. **Projection** computes the set of constraints $\Sigma_{MAC}$ shown in Table 7. For example, **Projection** returns the third constraint in Table 7 because $G(\Sigma_{APO})$ has a path from the node labeled with mo:Label to the node labeled with ¬(≥1 foaf:name) (and, hence, $G^*(\Sigma_{APO})$ has an arc between these two nodes), and likewise for the forth constraint. □

**Table 7.** Constraints of ontology ***MAC*** (unabbreviated form).

| | Constraint | Informal specification |
|---|---|---|
| 1 | (≥1 foaf:name$^{-}$) ⊑ xsd:string | The range of foaf:name is xsd:string |
| 2 | mo:SoloMusicArtist ⊑ mo:MusicArtist<br>mo:MusicGroup ⊑ mo:MusicArtist | mo:SoloMusicArtist is a subset of mo:MusicArtist<br>mo:MusicGroup is a subset of mo:MusicArtist |
| 3 | mo:Label ⊑ ¬(≥1 foaf:name) | $G(\Sigma_{APO})$ has a path from the node labeled with mo:Label to the node labeled with ¬(≥1 foaf:name) (which indicates that a label has no name) |
| 4 | mo:Label ⊑ ¬mo:SoloMusicArtist | $G(\Sigma_{APO})$ has a path from the node labeled with mo:Label to the node labeled with ¬mo:SoloMusicArtist (which indicates that mo:SoloMusicArtist and mo:Label are disjoint) |

## 5.3 Implementation of Intersection

Let $O_k = (V_k,\Sigma_k)$, for $k=1,2$, be two lightweight ontologies. Recall that the intersection of $O_1$ and $O_2$ is the ontology $O_N = (V_N,\Sigma_N)$, where $V_N = V_1 \cap V_2$ and $\Sigma_N = \tau[\Sigma_1] \cap \tau[\Sigma_2]$.

Procedure **Intersection** computes $\Gamma_N$ so that $\tau[\Gamma_N] = \tau[\Sigma_N]$. That is, a lightweight inclusion is a logical consequence of $\Gamma_N$ iff it is a logical consequence of $\Sigma_k$, for $k=1,2$.

We proceed to discuss the major decisions that lead to the **Intersection** procedure. First recall from Theorem 1 that a lightweight inclusion $e \sqsubseteq f$ is a logical consequence of $\Sigma_k$ iff there are nodes $M$ and $N$ of $G(\Sigma_k,\Omega)$, with $\Omega=\{e,f\}$, such that

(i) The node of $G(\Sigma_k,\Omega)$ labeled with $e$ is a $\bot$-node, or
(ii) The node of $G(\Sigma_k,\Omega)$ labeled with $f$ is a $\top$-node, or
(iii) There is a path in $G(\Sigma_k,\Omega)$, possibly with length 0, from the node labeled with $e$ to the node labeled with $f$.

Therefore, we must construct $\Gamma_N$ so that $e \sqsubseteq f$ is a logical consequence of $\Gamma_N$ iff $e \sqsubseteq f$ satisfies the above conditions with respect to $\Sigma_k$, for $k=1,2$. However, a direct application of Theorem 1 depends on the inclusion $e \sqsubseteq f$ being tested (since the theorem depends on the constraint graph $G(\Sigma,\Omega)$, with $\Omega$=□□□ □□). We argue that we can simplify the application of Theorem 1 in the context of the intersection operation, if we define a set of concept descriptions as follows.

**Definition 10**: Let $\Sigma_1$ and $\Sigma_2$ be two sets of lightweight inclusions. The *closure of $\Sigma_1$ and $\Sigma_2$ with respect to each-other* is the set $\Delta$ of concept descriptions defined so that a concept description $e$ is in $\Delta$ iff, for $k=1,2$, $e$ occurs in an inclusion of $\Sigma_k$ but not in an inclusion of $\Sigma_{k+1}$ (sum is module 2). □

Then, $G(\Sigma_1,\Delta)$ and $G(\Sigma_2,\Delta)$ satisfy the following property.

**Proposition 4**: Let $\Sigma_1$ and $\Sigma_2$ be two sets of lightweight inclusions and $\Delta$ be the closure of $\Sigma_1$ and $\Sigma_2$ with respect to each other. Then, for $k=1,2$, any lightweight inclusion $e \sqsubseteq f$ in $\Sigma_{k+1}$ (sum is module 2) is a logical consequence of $\Sigma_k$ iff there are nodes $M$ and $N$ of $G(\Sigma_k,\Delta)$ such that

(i) The node of $G(\Sigma_k,\Delta)$ labeled with $e$ is a $\bot$-node, or
(ii) The node of $G(\Sigma_k,\Delta)$ labeled with $f$ is a $\top$-node, or
(iii) There is a path in $G(\Sigma_k,\Delta)$, possibly with length 0, from the node labeled with $e$ to the node labeled with $f$. □

The final case analysis to compute the intersection operation is summarized in Table 8 and results in a set of lightweight inclusions (Column C of Table 8). Step (3), the core of the **Intersection** procedure, directly captures such case analysis. We decided to create a set of lightweight inclusions, rather than a constraint graph, just to clarify the decisions behind the **Intersection** procedure. The actual implementation is optimized and avoids this intermediate step.

Note that we need not consider $\top$-nodes of $G(\Sigma_1,\Delta)$ (or of $G(\Sigma_2,\Delta)$). Indeed, by Proposition 1, there is a $\top$-node of $G(\Sigma_1,\Delta)$ labeled with $f$ iff there is a $\bot$-node labeled with $\bar{f}$. Furthermore, there is a path, possibly with length 0, from a node labeled with $e$ to a node labeled with $f$ iff there is a path, possibly with length 0, from a node labeled with $\bar{f}$ to a node labeled with $\bar{e}$. Therefore, Cases 4, 5 and 6 of Table 8 respectively reduce to Cases 2, 1 and 3 of Table 8.

**Table 8.** Case analysis for the intersection operation.

| Case | (A) Condition on $G(\Sigma_1,\Delta)^{1,2}$ | (B) Condition on $G(\Sigma_2,\Delta)$ | (C) Inclusion in $\Sigma_3$ |
|---|---|---|---|
| 1 | there is a $\bot$-node labeled with $e$ | there is a $\bot$-node labeled with $e$ | $e \sqsubseteq \bot$ |
| 2 | | there is a $\top$-node labeled with $f$ | $e \sqsubseteq f$ |
| 3 | | there is a path, possibly with length 0, from a node labeled with $e$ to a node labeled with $f$ | $e \sqsubseteq f$ |
| 4 | there is a $\top$-node labeled with $f$ | there is a $\bot$-node labeled with $e$ | $e \sqsubseteq f$ |
| 5 | | there is a $\top$-node labeled with $f$ | $\top \sqsubseteq f$ |
| 6 | | there is a path, possibly with length 0, from a node labeled with $e$ to a node labeled with $f$ | $e \sqsubseteq f$ |
| 7 | there is a path, possibly with length 0, from a node labeled with $e$ to a node labeled with $f$ | there is a $\bot$-node labeled with $e$ | $e \sqsubseteq f$ |
| 8 | | there is a $\top$-node labeled with $f$ | $e \sqsubseteq f$ |
| 9 | | there is a path, possibly with length 0, from a node labeled with $e$ to a node labeled with $f$ | $e \sqsubseteq f$ |

Notes (see Definition 2):
(1) $e$ is an atomic concept or an at-least restriction.
(2) $f$ is the bottom concept $\bot$, an atomic concept, a lightweight at-least restriction, a negated atomic concept or a negated at-least restrictions.

Procedure **Intersection** works as follows:

(1) Construct the closure $\Delta$ of $\Sigma_1$ and $\Sigma_2$ with respect to each other.
(2) Construct $G(\Sigma_1,\Delta)$ and $G(\Sigma_2,\Delta)$, the tagged constraint graphs for $\Sigma_1$ and $\Delta$ and $\Sigma_2$ and $\Delta$, respectively.
(3) Construct a set of constraints $\Sigma_3$ as follows (see Table 8):
  (a) Initialize $\Sigma_3$ to be the empty set.
  (b) For each node $M$ of $G(\Sigma_1,\Delta)$ tagged with "$\bot$-node" and
      labeled only with atomic concepts and at-least restrictions,
    for each label $e$ of $M$, do:
    (i) If $e$ also labels a node of $G(\Sigma_2,\Delta)$ tagged with "$\bot$-node", then
        add $e \sqsubseteq \bot$ to $\Sigma_3$.
    (ii) For each node $K$ of $G(\Sigma_2,\Delta)$ tagged with "$\top$-node",
        for each label $f$ of $K$,
        add $e \sqsubseteq f$ to $\Sigma_3$.
    (iii) For each path of $G(\Sigma_2,\Delta)$, possibly with length 0, from a node labeled with $e$ to a node labeled with $f$,
        add $e \sqsubseteq f$ to $\Sigma_3$.
  (c) For each node $M$ of $G(\Sigma_1,\Delta)$ not tagged with "$\bot$-node" and
      labeled only with atomic concepts and at-least restrictions,
    for each path in $G(\Sigma_1,\Delta)$, possibly with length 0, from $M$ to a node $N$,
    for each label $e$ of $M$,
    for each label $f$ of $N$ ($f \neq e$, if $M=N$), do:
    (i) If $e$ also labels a node of $G(\Sigma_2,\Delta)$ tagged with "$\bot$-node", then
        add $e \sqsubseteq f$ to $\Sigma_3$.
    (ii) If $f$ also labels a node of $G(\Sigma_2,\Delta)$ tagged with "$\top$-node", then
        add $e \sqsubseteq f$ to $\Sigma_3$.

(iii) If there is a path in $G(\Sigma_2,\Delta)$, possibly with length 0, from a node labeled with $e$ to a node labeled with $f$, then add $e \sqsubseteq f$ to $\Sigma_3$.

(4) Call **MinimizeContraints** with $\Sigma_3$ to generate $\Gamma_N$.

(5) Return $O_N = (V_1 \cap V_2, \Gamma_N)$.

**Theorem 4** (Correctness of **Intersection**): Let $O_1 = (V_1,\Sigma_1)$ and $O_2= (V_2,\Sigma_2)$ be two sets of lightweight ontologies. Let $\Delta$ be the closure of $\Sigma_1$ and $\Sigma_2$ with respect to each other. Let $O_N =(V_1 \cap V_2,\Gamma_N)$ be the ontology that **Intersection** returns for $O_1$ and $O_2$. Let $e \sqsubseteq f$ be a lightweight inclusion. Then, $\Gamma_N \vDash e \sqsubseteq f$ iff $\Sigma_1 \vDash e \sqsubseteq f$ and $\Sigma_2 \vDash e \sqsubseteq f$. □

### 5.4 A Discussion on Difference

The problem of creating a procedure to compute the difference between two ontologies, $O_1=(V_1,\Sigma_1)$ and $O_2=(V_2,\Sigma_2)$, lies in that it might not be possible to obtain a finite set of inclusions $\Delta_N$ in such a way that

(1) $\tau[\Delta_N] = \tau[\Sigma_1] - \tau[\Sigma_2]$

This invalidates the effort to create a procedure to obtain a finite set of inclusions $\Delta_N$ satisfying (1), along the lines of those exhibited in Sections 5.2 and 5.3. This remark in fact puts in doubt the usefulness of a (generic) difference operation.

For example, consider the following two sets of inclusions:

(2) $\Sigma_1 = \{ e \sqsubseteq g, g \sqsubseteq f \}$

(3) $\Sigma_2 = \{ e \sqsubseteq f \}$

Then, ignoring tautologies when computing $\tau[\Sigma_j], j=1,2$, we have:

(4) $\tau[\Sigma_1] = \{ e \sqsubseteq g, g \sqsubseteq f, e \sqsubseteq f \}$

(5) $\tau[\Sigma_2] = \{ e \sqsubseteq f \}$

(6) $\Delta_N = \tau[\Sigma_1] - \tau[\Sigma_2] = \{ e \sqsubseteq g, g \sqsubseteq f \} = \Sigma_1$

But this definition of $\Delta_N$ is not satisfactory, since we have

(7) $\tau[\Delta_N] = \tau[\Sigma_1] = \{ e \sqsubseteq g, g \sqsubseteq f, e \sqsubseteq f \}$

That is, to compute the difference $\Delta_N = \tau[\Sigma_1] - \tau[\Sigma_2]$, we remove "$e \sqsubseteq f$" from $\tau[\Sigma_1]$, only to get "$e \sqsubseteq f$" back by logical implication from $\Delta_N$. In fact, in this rather obvious example, we cannot obtain a set of inclusions $\Delta_N$ such that $\tau[\Delta_N] = \tau[\Sigma_1] - \tau[\Sigma_2]$. Indeed, since the set of inclusions must not logically imply "$e \sqsubseteq f$", the only candidates are:

(8) $\Delta_1 = \{ e \sqsubseteq g \}$

(9) $\Delta_2 = \{ g \sqsubseteq f \}$

In both cases, we have that (again ignoring tautologies when computing $\tau[\Delta_k]$, $k=1,2$):

(10) $\tau[\Delta_k] = \Delta_k \subset \tau[\Sigma_1] - \tau[\Sigma_2]$

A non-deterministic procedure **Difference** to compute a subset of $\tau[\Sigma_1] - \tau[\Sigma_2]$ would be:

(1) Construct the closure $\Delta$ of $\Sigma_1$ and $\Sigma_2$ with respect to each other.

(2) Construct $G(\Sigma_1,\Delta)$ and $G(\Sigma_2,\Delta)$, the tagged constraint graphs for $\Sigma_1$ and $\Delta$ and $\Sigma_2$ and $\Delta$, respectively.
(3) Construct a constraint graph $H$ as follows:
  (a) Initialize $H = G(\Sigma_1,\Delta)$.
  (b) For each node $M$ of $G(\Sigma_2,\Delta)$ tagged with "⊥-node"
      for each label $e$ of $M$, do:
        for each node $K$ of $H$ such that $K$ is labeled with $e$, do:
          Drop $K$ from $H$, or
          Drop from $H$ all arcs leaving $K$.
  (c) For each node $N$ of $G(\Sigma_2,\Delta)$ tagged with "⊤-node"
      for each label $f$ of $N$, do:
        for each node $L$ of $H$ such that $K$ is labeled with $f$, do:
          Drop $L$ from $H$, or
          Drop from $H$ all arcs entering $L$.
  (d) For each path in $G(\Sigma_2,\Delta)$, possibly with length 0, from node $M$ to node $N$,
      for each label $e$ of $M$,
        for each label $f$ of $N$ ($f \neq e$, if $M$=$N$), do:
        (i) If $e$ labels a node $K$ of $G(\Sigma_1,\Delta)$ tagged with "⊥-node", then
            Drop $K$ from $H$, or
            Drop from $H$ all arcs leaving $K$.
        (ii) If $f$ labels a node $L$ of $G(\Sigma_1,\Delta)$ tagged with "⊤-node", then
            Drop $L$ from $H$, or
            Drop from $H$ all arcs entering $L$
        (iii) If there is a path in $G(\Sigma_1,\Delta)$, possibly with length 0, from a node $K$ labeled with $e$ to a node $L$ labeled with $f$, then
            Drop $K$ from $H$, or
            Drop $L$ from $H$, or
            Drop arcs from $H$ until there are no paths between $K$ and $L$.
(4) Call **MinimizeGraph** with $H$ to generate $H'$.
(5) Call **GenerateContraints** with $H'$ to generate $\Gamma_D$.
(6) Return $O_D$=$(V_1,\Gamma_D)$.

**Theorem 5** (Correctness of **Difference**): Let $O_1 = (V_1,\Sigma_1)$ and $O_2 = (V_2,\Sigma_2)$ be two sets of lightweight ontologies. Let $\Delta$ be the closure of $\Sigma_1$ and $\Sigma_2$ with respect to each other. Let $O_D = (V_1, \Gamma_D)$ be the ontology that **Difference** returns for $O_1$ and $O_2$. Then, for any lightweight inclusion $e \sqsubseteq f$, if $\Gamma_D \vDash e \sqsubseteq f$, then $\Sigma_1 \vDash e \sqsubseteq f$ but not $\Sigma_2 \vDash e \sqsubseteq f$. □

## 6 A Protégé Plugin Implementation

The **OntologyManagerTab**, presented in this section, offers the ontology operation described in previous sections, integrated with traditional ontology management features. The tool was developed in Java as a tab plug-in over Protégé 3.4.8 (the implementation might require minor modifications to work with other versions of Protégé).

Despite the fact that **OntologyManagerTab** was developed as a Protégé plug-in, it works in a completely independent manner from the main framework, using Protégé only as a Graphical User Interface (GUI) enclosure. In other words, all the functionalities provided by OntologyManagerTab do not rely on any of the Protégé libraries, making the tool easier to adapt as a plug-in for any other frame-work or as a stand-alone software.

**OntologyManagerTab** normalizes and loads OWL ontologies to be operated upon and saves the resulting ontology as an OWL file. The plugin uses a two-column table to represent the constraints of an ontology, where:

- each expression that occurs in a constraint of the ontology appears at least once in the first column;
- each constraint $e \sqsubseteq f$ of the ontology appears in a separate line of the table, where $e$ appears in the first column and $f$ in the second;
- if any bottom nodes are found there will be a row with its name in Column 1 and $\bot$ in Column 2.

To use the plugin you will need to run Protégé and open or create a new project as shown bellow on figures 4 and 5, where we chose to open the newspaper example.

Afterwards we need to enable our plugin "OntologyManagerTab" to run as a "Tab Widget". We follow the path Project->Configure select our plugin and click "OK" which will open the OntolgyManagerTab as shown on Figures 6, 7 and 8.

To load an ontology you need to either click on the "Load Ontology 1" or "Load Ontology 2" button and a browse window will appear as shown on Figure 9. In this example we chose to load the FOAF ontology, the plugin will normalize the ontology and load the result to a restriction graph in memory. We can see the tab feedback on Figure 10, wich shows the Graph axioms, and the normalized result on Figure 11. When loading an ontology the default visualization shows each element full IRI, to help the user with a better and easier interface we have a button that shows and hides this, in Figure 10 all IRIs have been hidden.

Aside from the projection all other operations are calculated over two ontologies. The projection option loads into the second table all the classes and properties from the ontology loaded in the first so that the user can choose over which nodes the operation will be done, as shown in Figure 12 and 13. Any other operation will request that a second ontology be loaded as show in Figure 14.

The OntologyManagerTab takes into account not only each node IRI but also the IRI of the original loaded ontology to run its operations, which allows a better matching when mapping correspondent nodes between ontologies. We now show an Intersection example between FOAF and MusicOntology in Figure 15.

It also provides a Graph minimization button, since we work with the transitive closures of each ontology it is very useful to extract the MEG for the obtained results. This minimization also finds equivalent cardinality restrictions and when possible collapse them into one.

OntologyManagerTab saves all resulting ontologies with "Normalized.owl" at its end, in Figure 16 and 17 we illustrate the saving of the ontology we obtained from the previous intersection as Test, the plugin automatically adds the "Normalized.owl" at its end resulting in the "TestNormalized.owl" file.

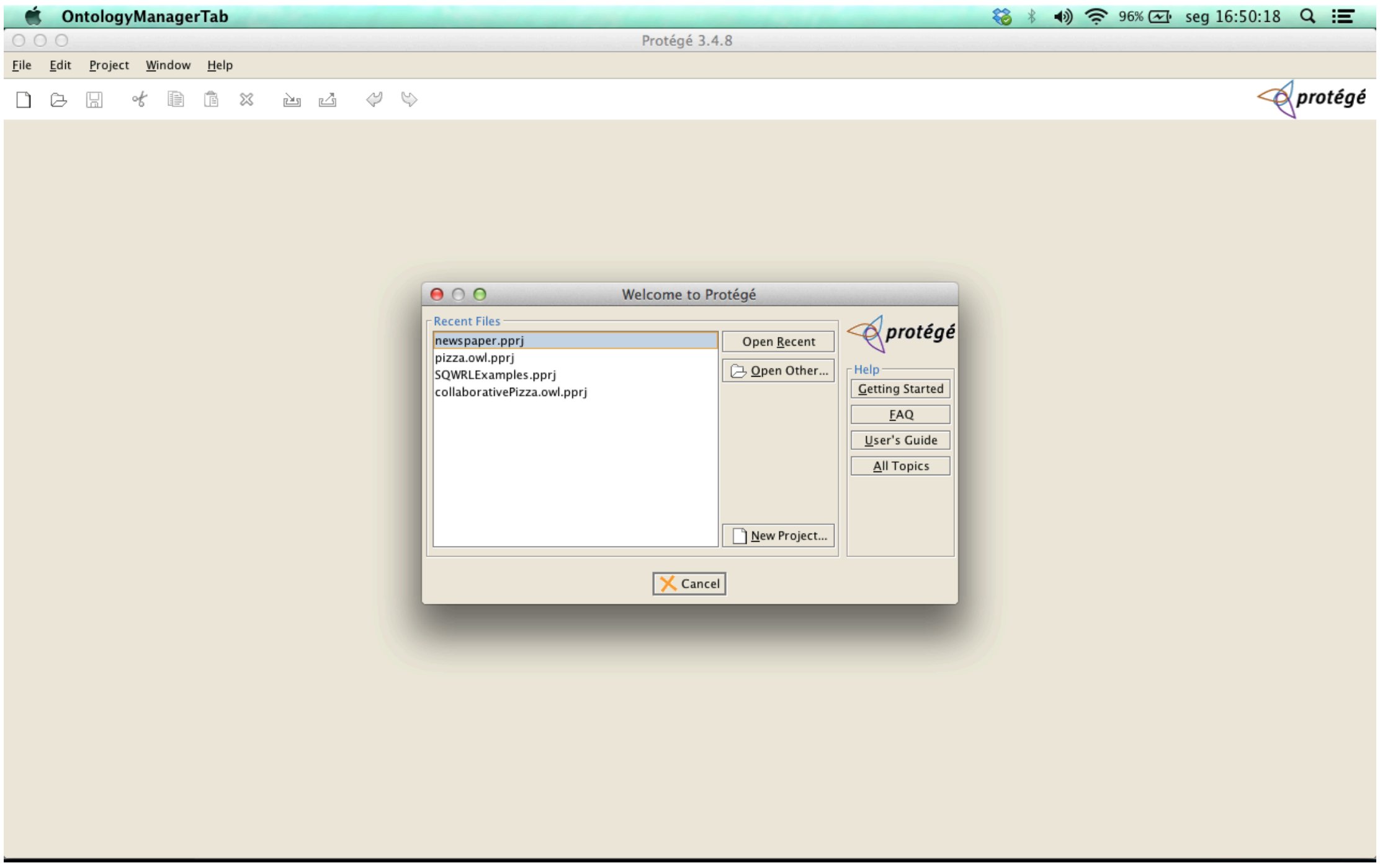

**Figure 4** – Running Protégé 3.4.8

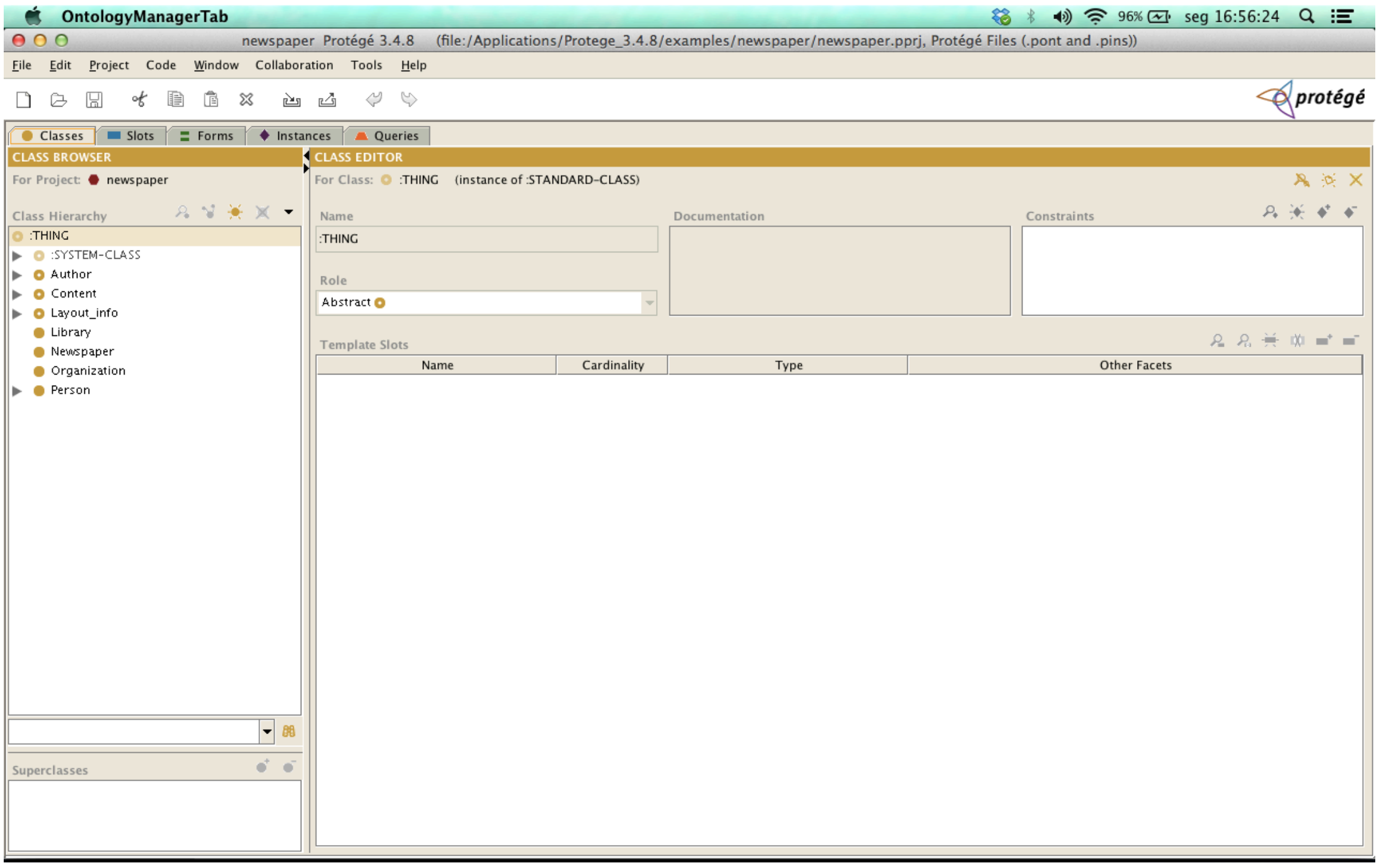

**Figure 5** – Opening newspaper example

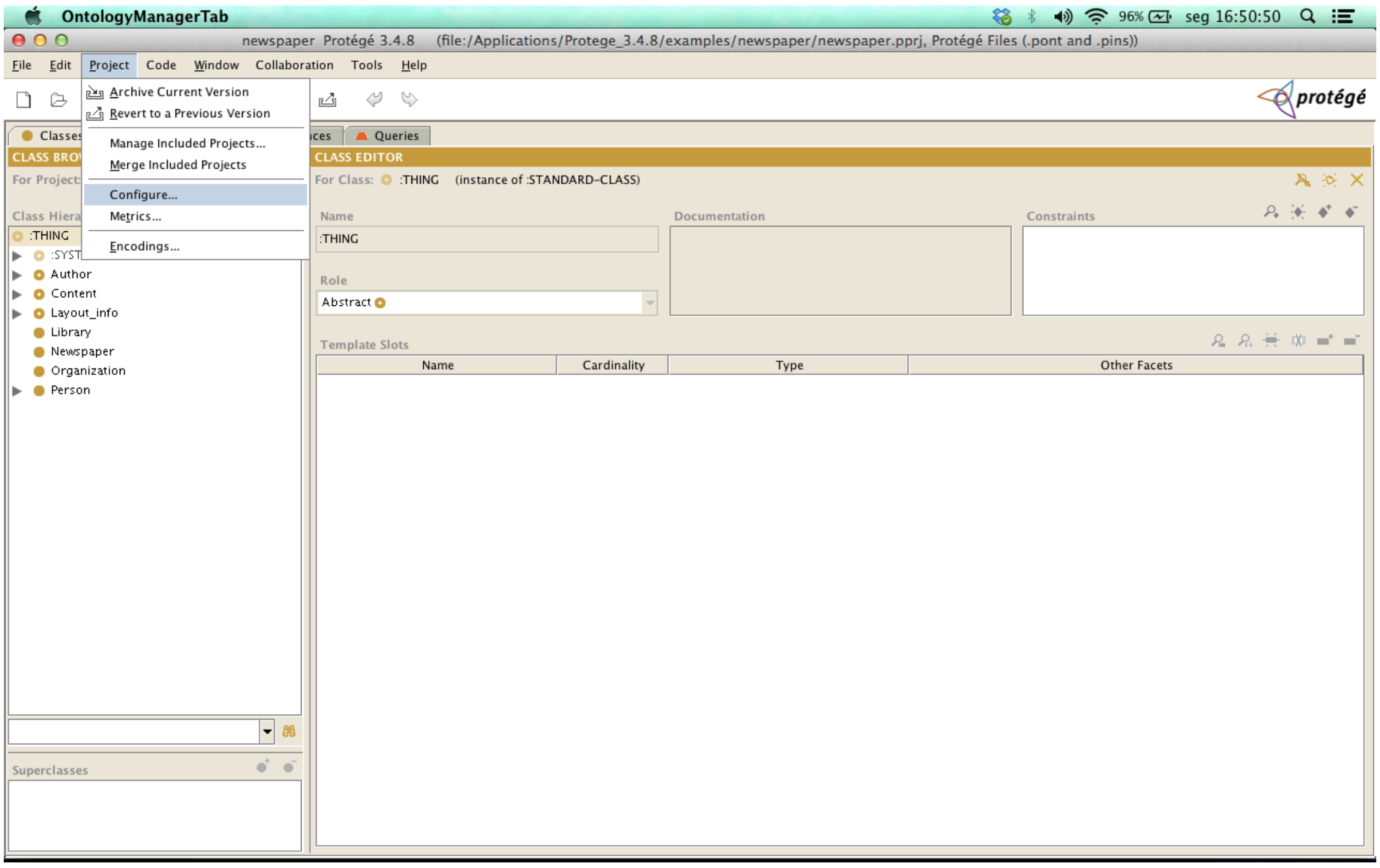

**Figure 6** – Configuring Protégé Widgets

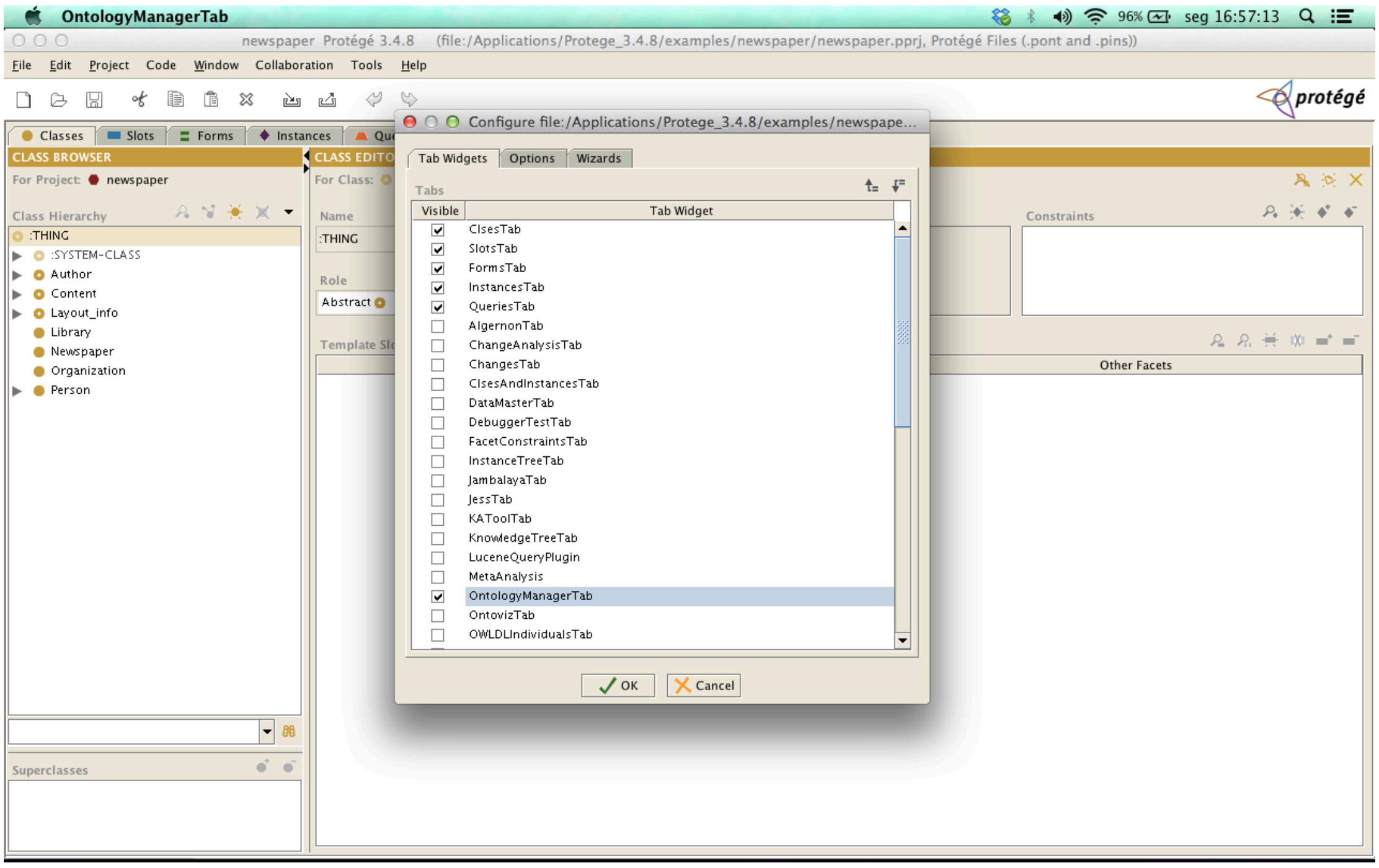

**Figure 7** – Selecting OntologyManagerTab

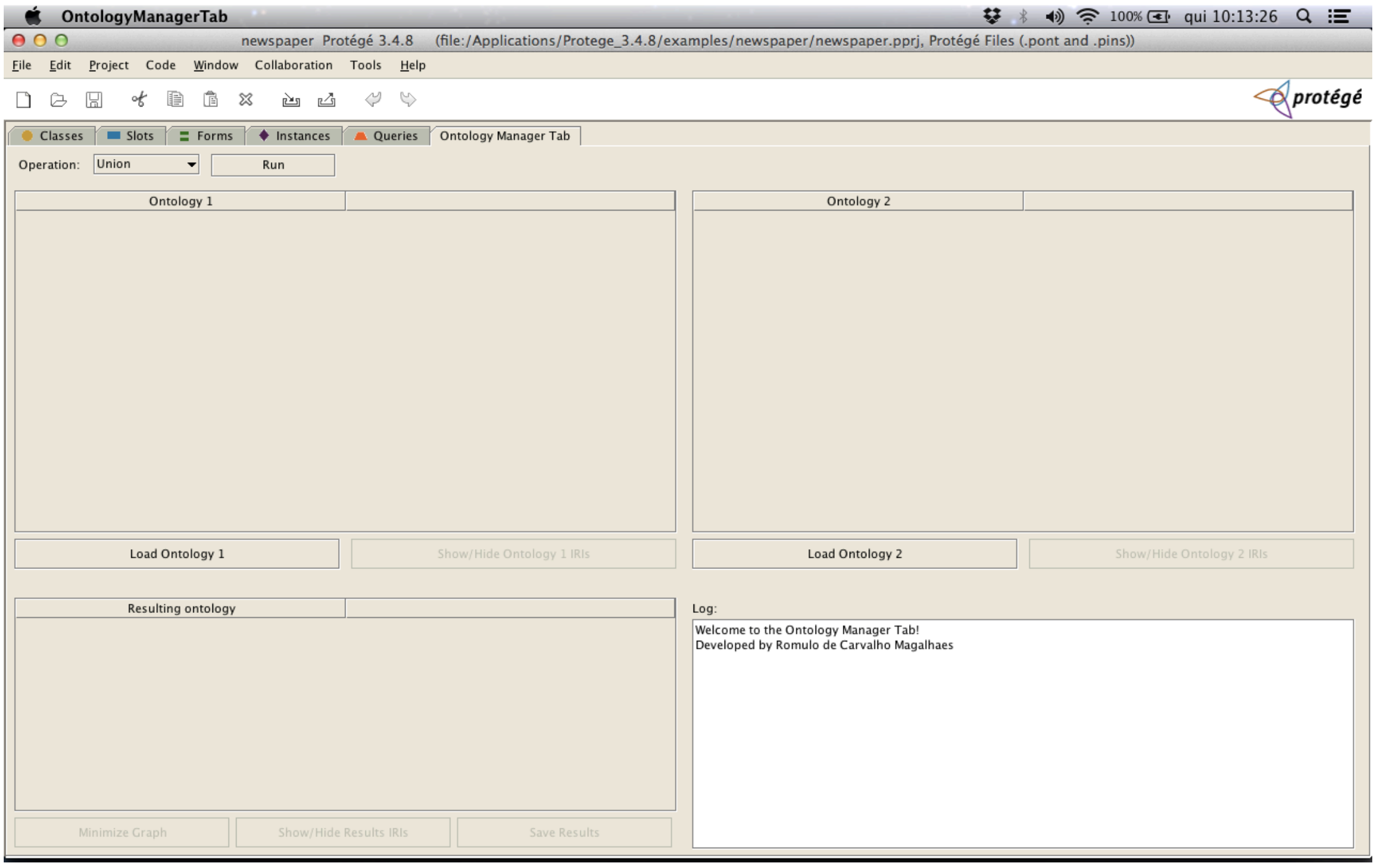

**Figure 8** – OntologyManagerTab

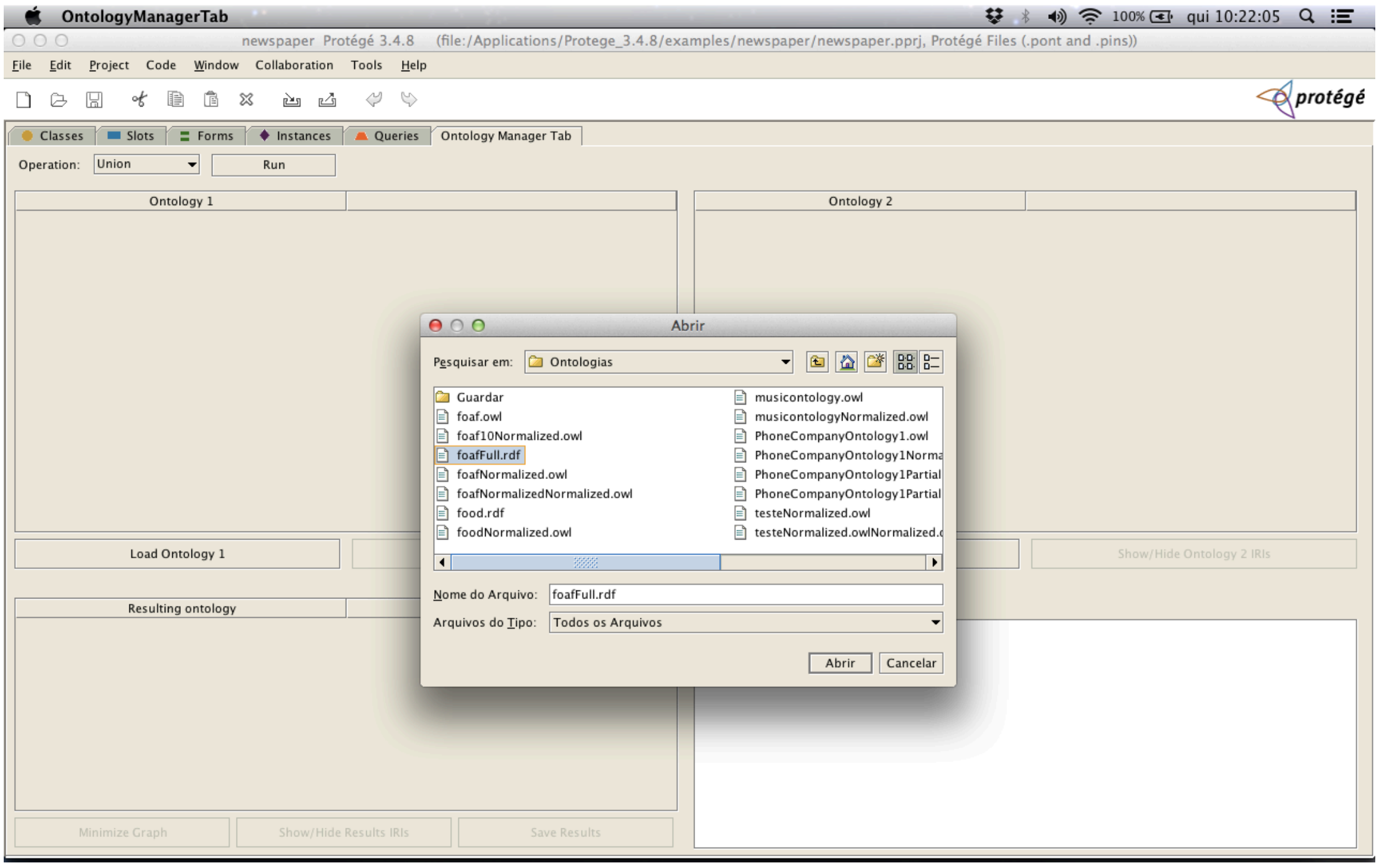

**Figure 9** – Browse Ontology Window

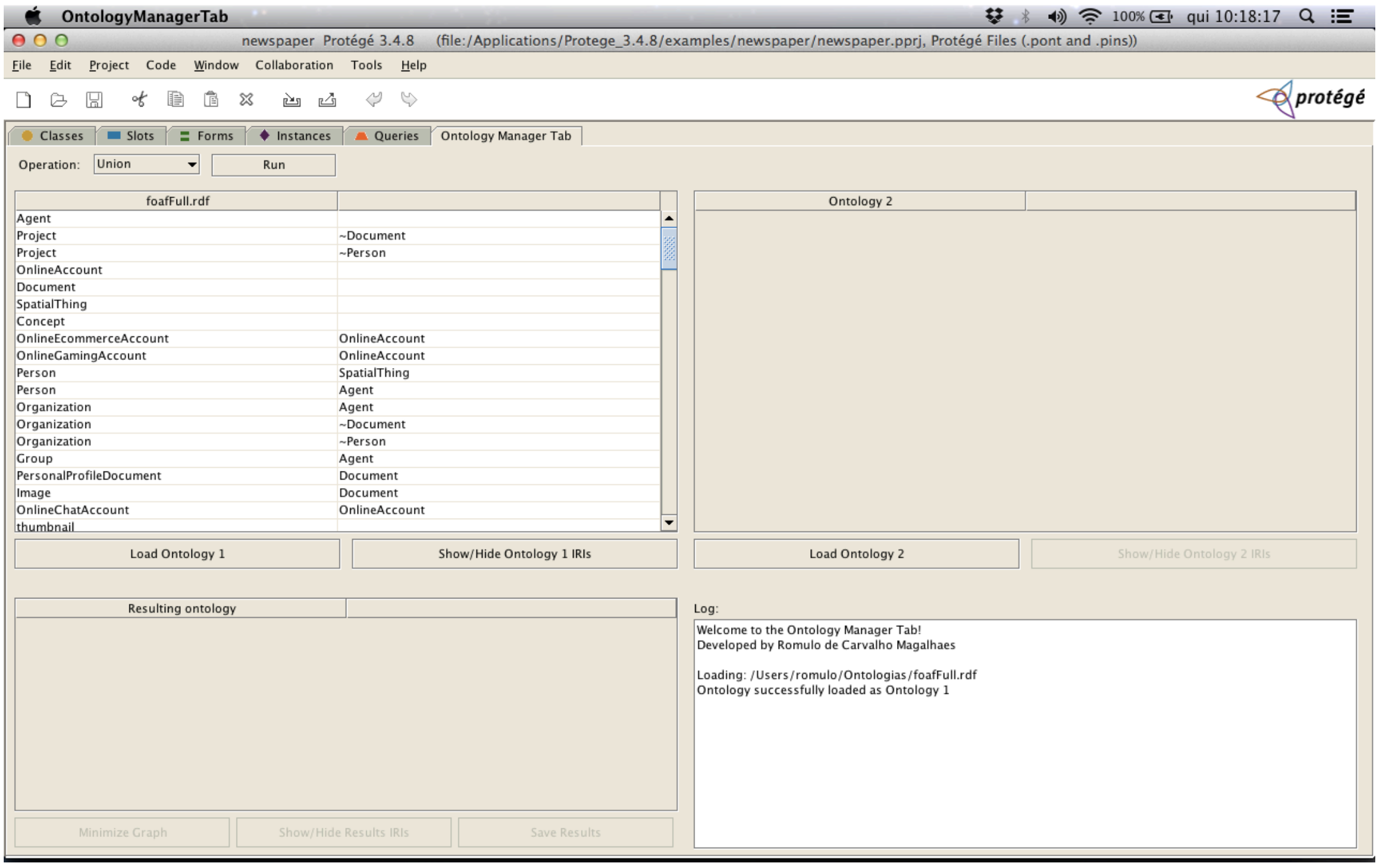

**Figure 10** – FOAF ontology loaded successful, with IRIs hidden

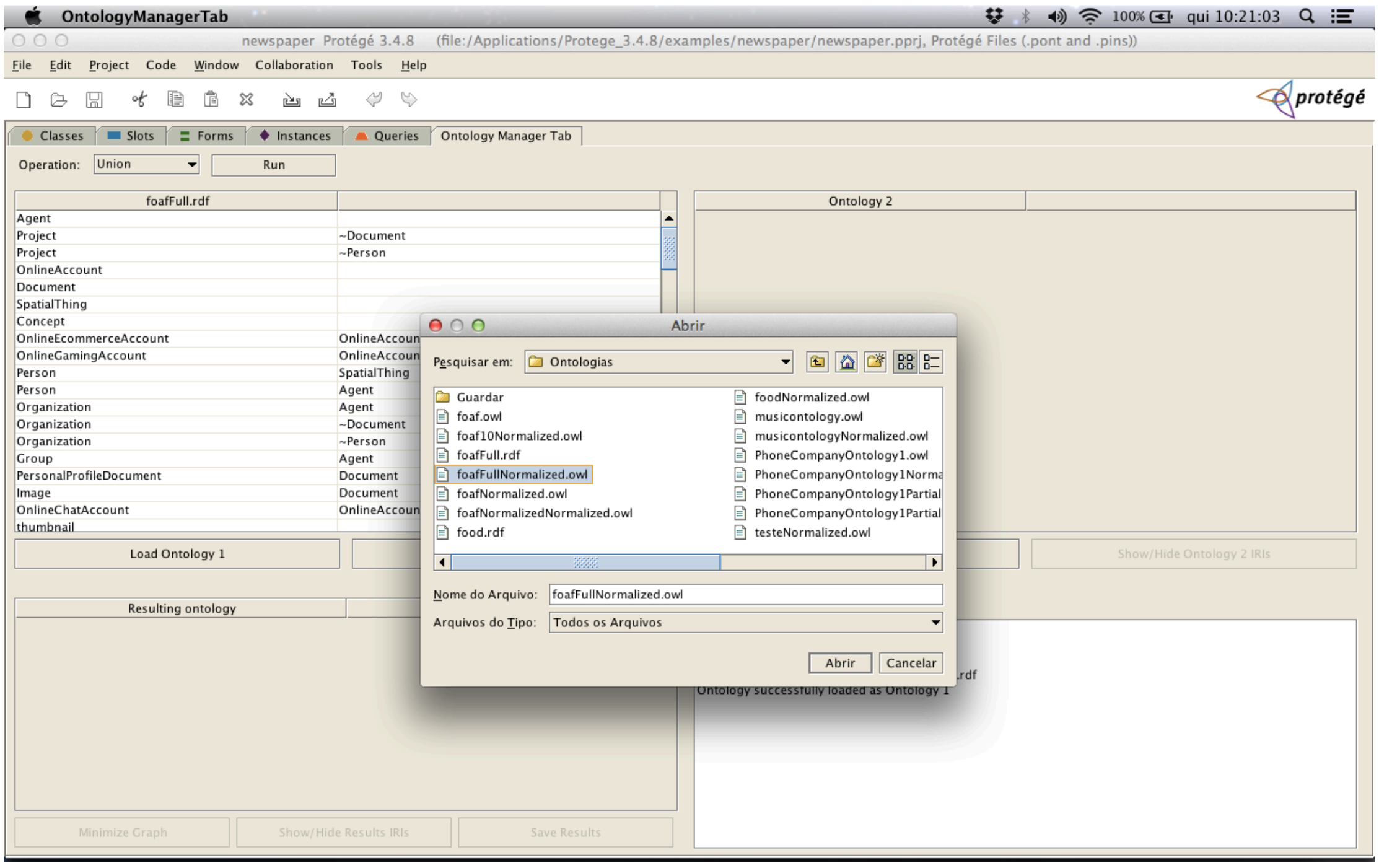

**Figure 11** – Showing resulting normalized file for foafFull.rdf

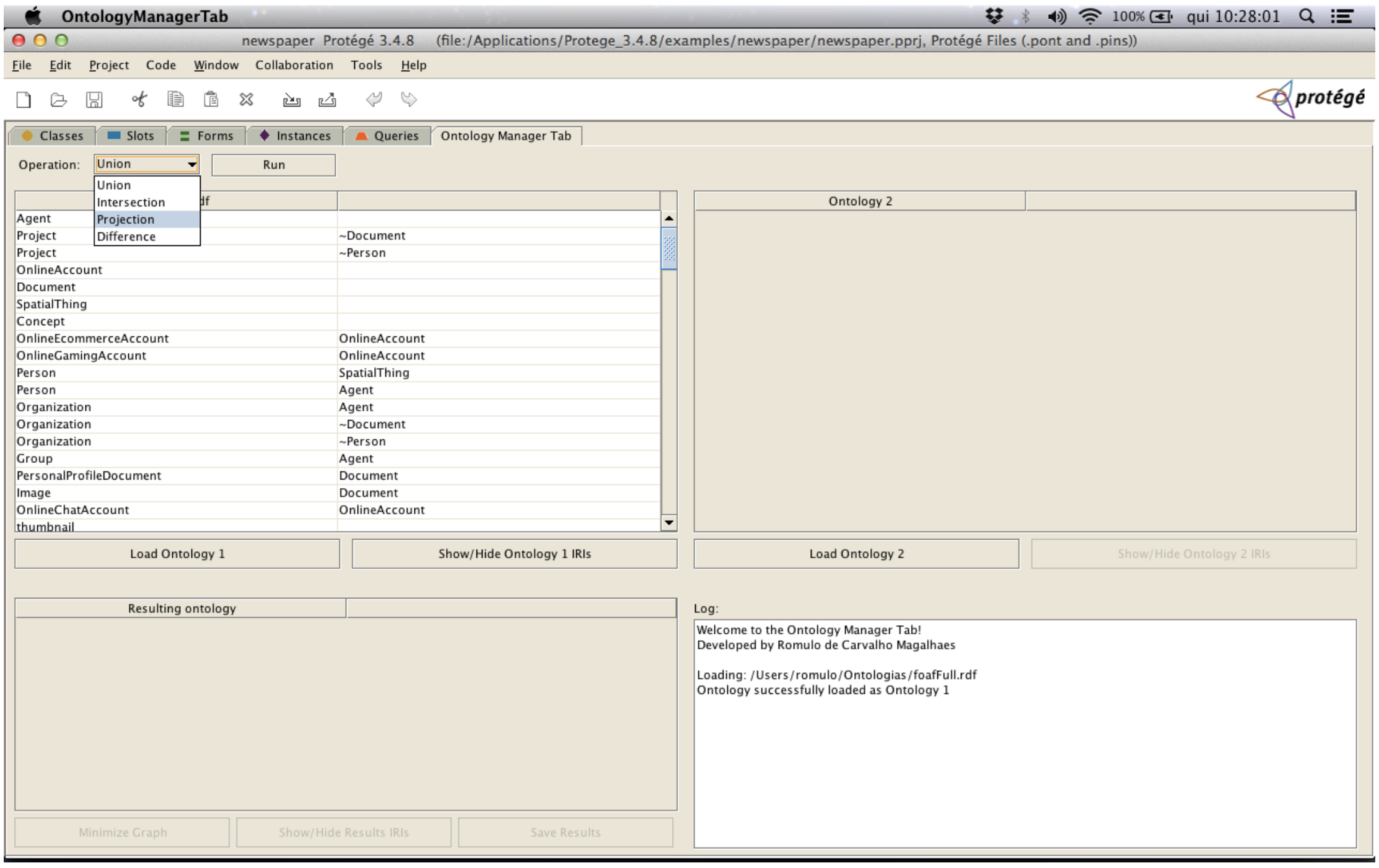

**Figure 12** – Selecting operation.

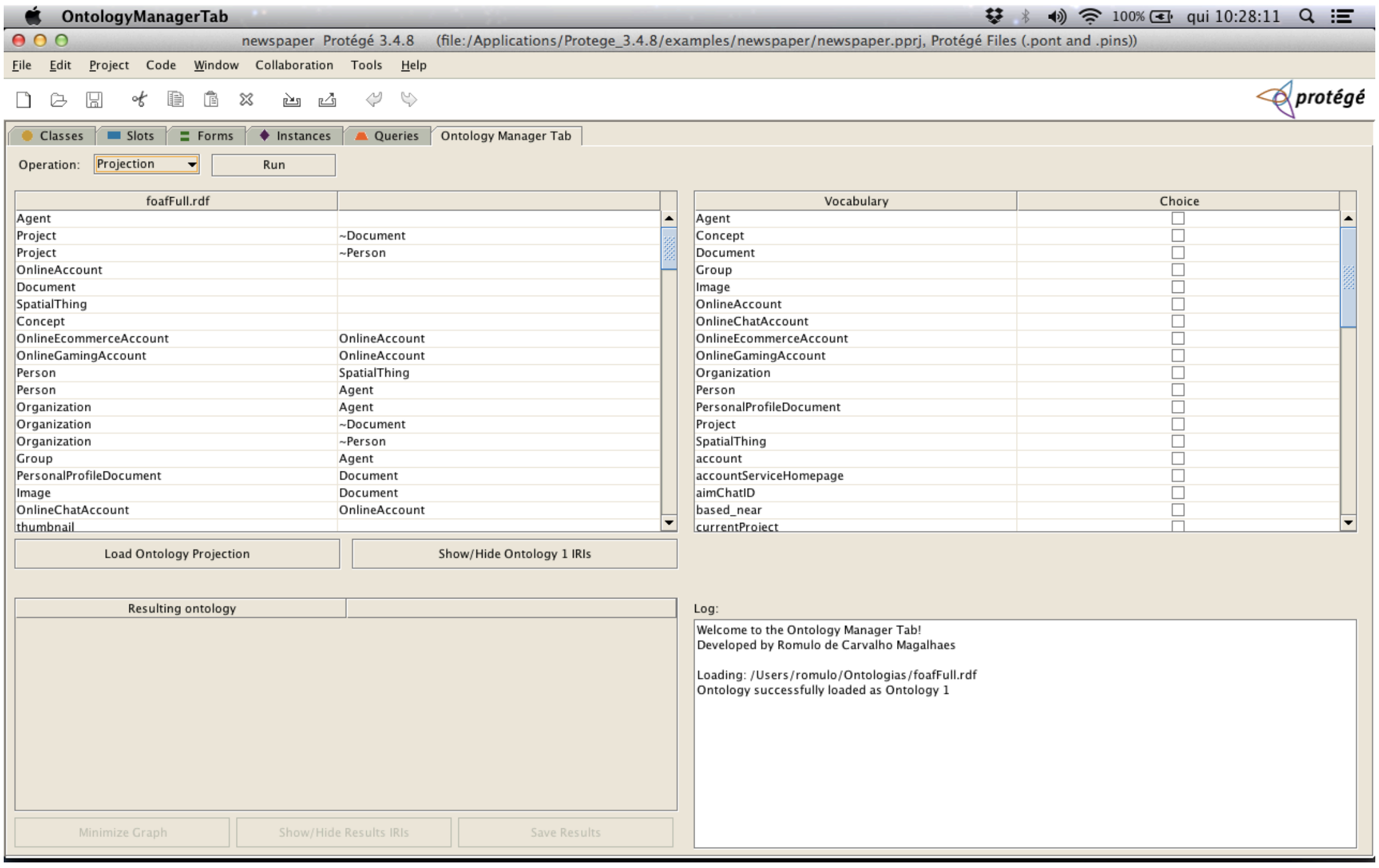

**Figure 13** – Showing projection selection for foafFull.rdf

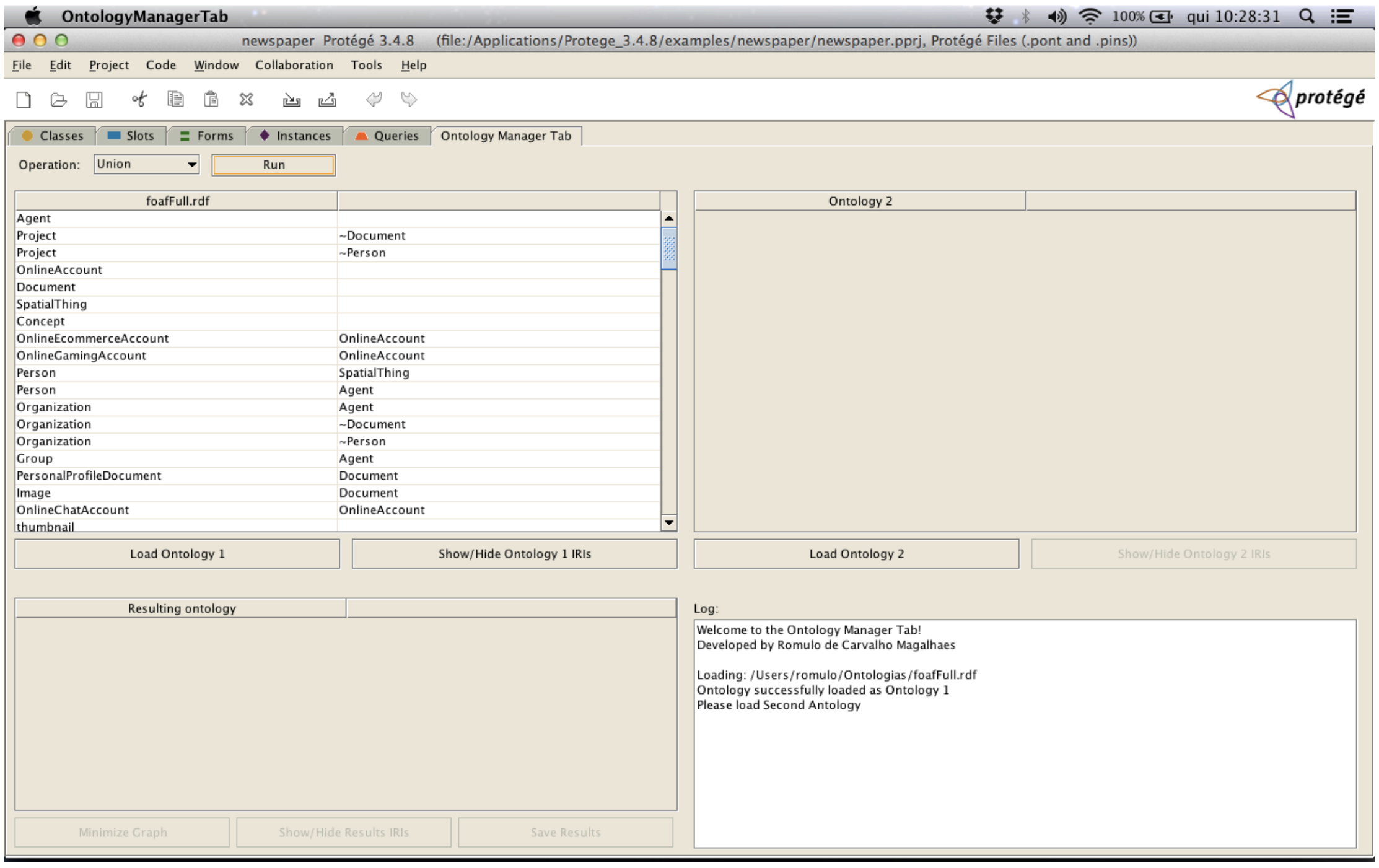

**Figure 14 –** OntologyManagerTab requesting second Ontology.

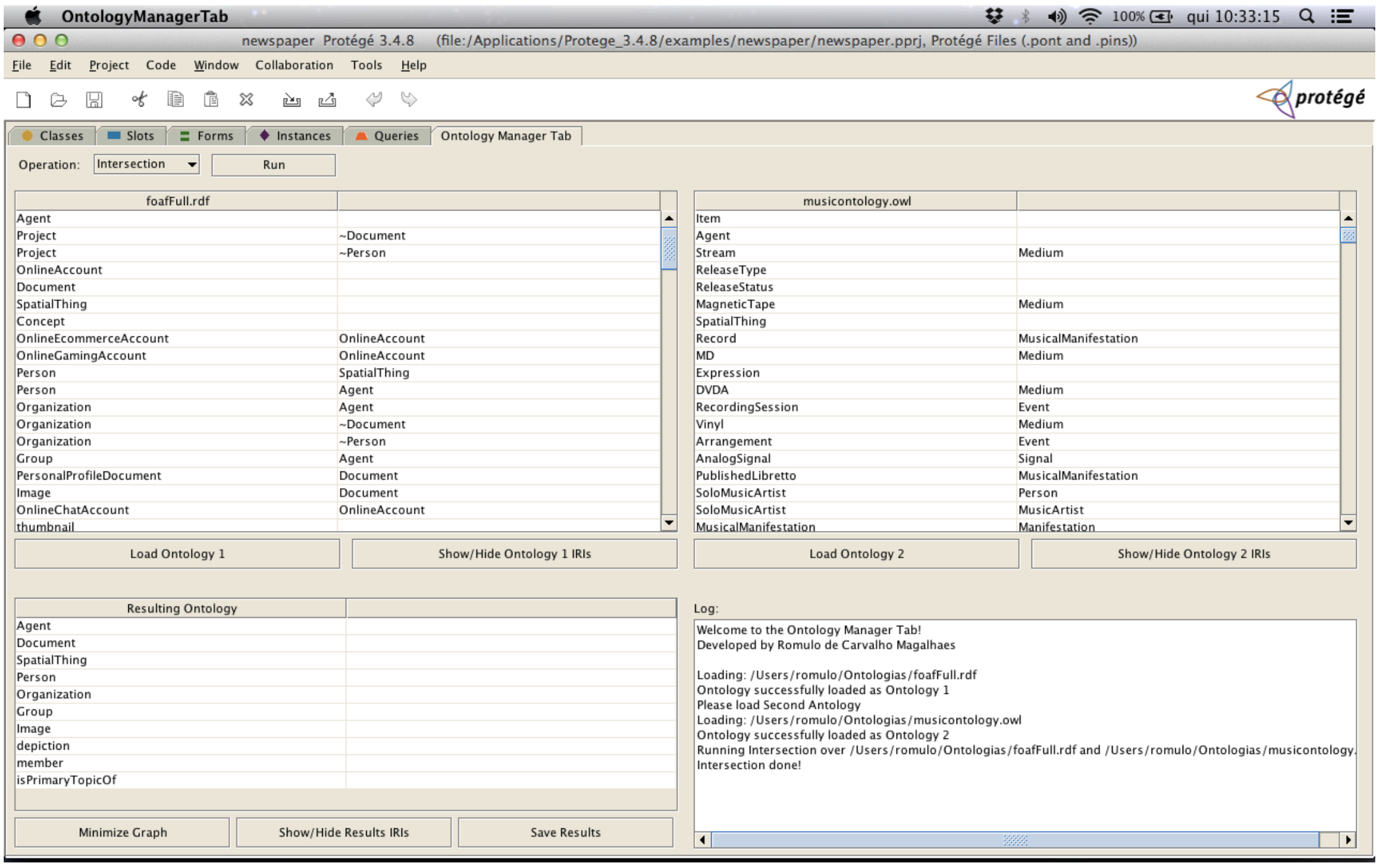

**Figure 15 –** Intersection between FOAF and MusicOntology.

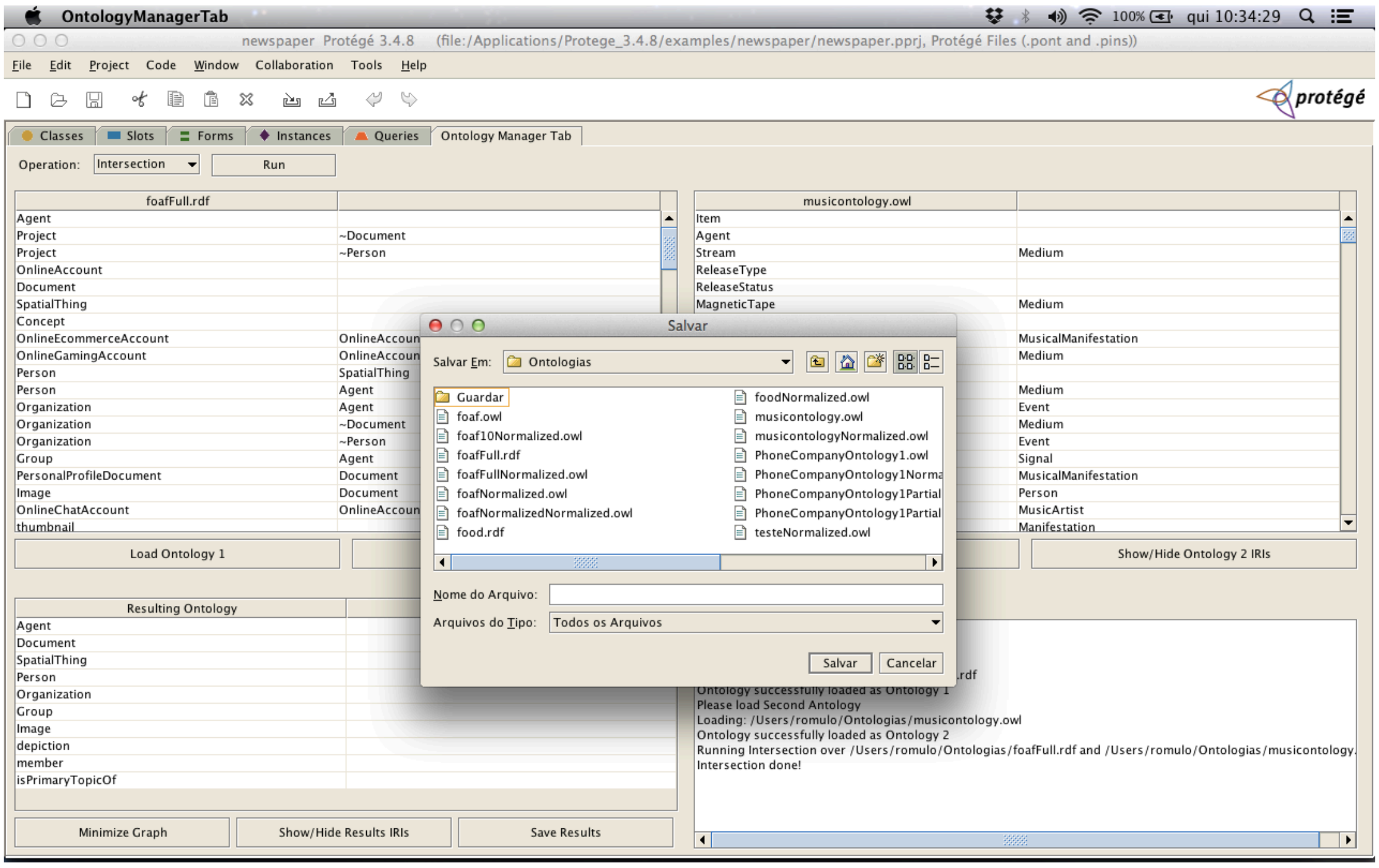

**Figure 16 –** Saving resulting ontology.

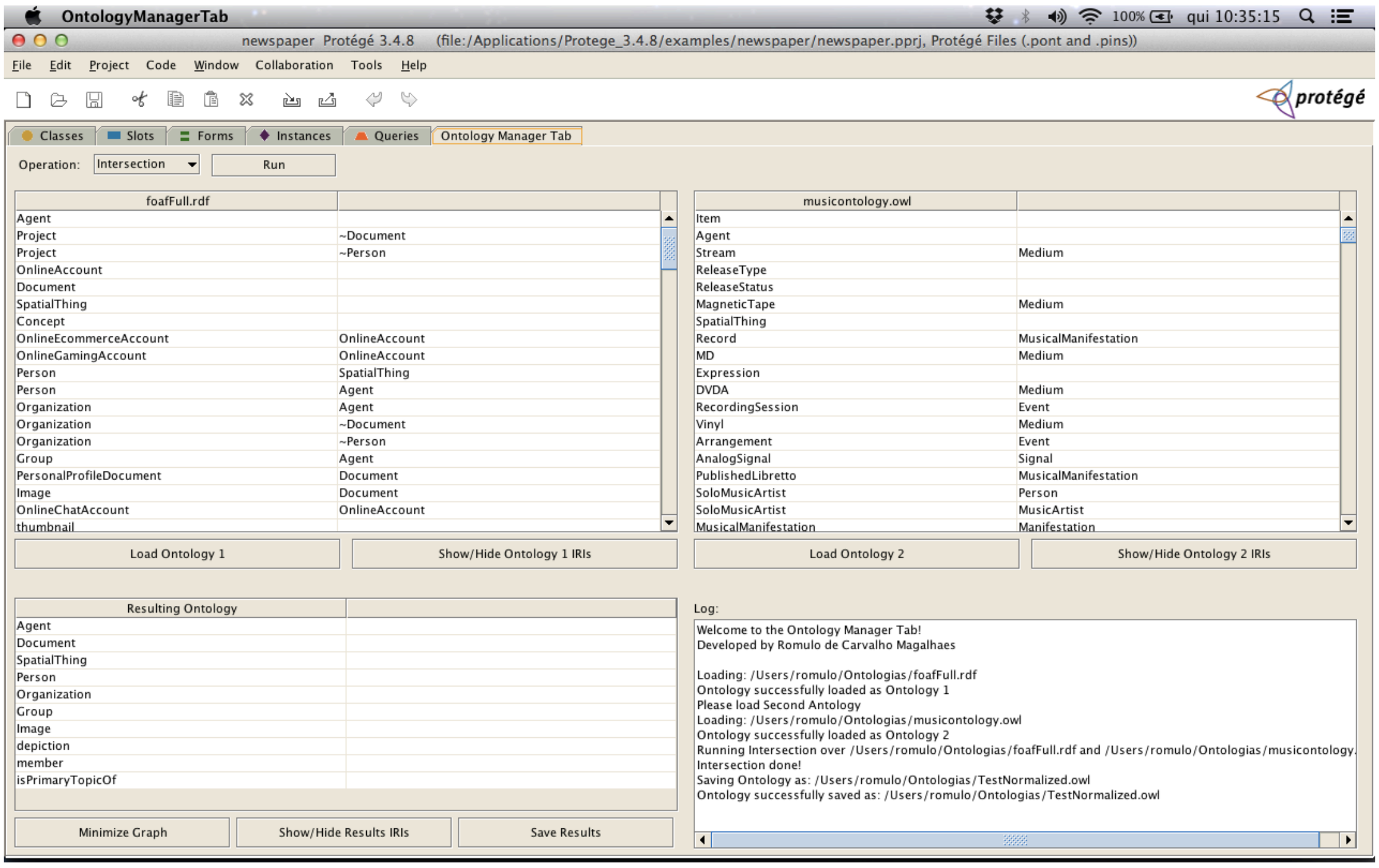

**Figure 17 –** Saving Ontology feedback.

# 7 Related Work

The results reported in the paper cover a topic – improving Linked Data design by constraint reuse – that is still neglected in the literature. The question of Linked Data semantics is not new, though. For example, recent investigation (Halpin and Haynes, 2010; Jaffrin et al., 2008; McCuster and McGuinness, 2010) in fact questions the correct use of owl:sameAs to inter-link datasets.

Jain et al. (2010) argues that the Linked Open Data (LoD) Cloud, in its current form, is only of limited value for furthering the Semantic Web vision. They discuss that the Linked Open Data Cloud can be transformed from "merely more data" to "semantically linked data" by overcoming problems such as lack of conceptual descriptions for the datasets, schema heterogeneity and absence of schema level links. Along this line, we advocated that the design of Linked Data sources must include constraints derived from those of the underlying ontologies.

We note that the problem we cover in this paper cannot be reduced to a question of ontology alignment in the context of Linked Data, addressed for example in (Leme et al., 2009; Prateek et al., 2010; Wang et al., 2011). Indeed, we stress that one of the problems we focus on refers to bootstrapping a new ontology (including its constraints) from one or more existing ontologies.

Some tools, such as Prompt (Noy and Musen, 2000) and ODEMerge (Ramos, 2001), allow the user to combine two or more ontologies in a semiautomatic or automatic way, respectively. Other tools, such as PromptDiff (Noy et al., 2004) and OntoDiff (Tury and Bieliková, 2006), deal with ontology change detection. However, these tools cannot capture changes in the semantics of the terms. The OntologyManagement tool described in (Pinheiro, 2013) circumvents this limitation since it is based on the operations described in this paper. Volz et al. (2003) proposes a tool that implements the projection operation by the creation of a database view resulting from query execution. However, this tool does not allow the generation of semantic information captured by the constraints that apply to the vocabulary terms.

Finally, previous work by the authors (Casanova et al., 2011) introduced the notion of open fragment, which is captured by the projection operation, whereas (Casanova et al., 2012b) considered the union and difference operations, the question of optimizing the representation of the resulting constraints and briefly described the OntologyManagement tool.

## 7 Conclusions

In this paper we argued that certain familiar ontology design problems could be profitably addressed by treating ontologies as theories and by defining a set of operations on ontologies. Such operations extend the idea of namespaces to take into account constraints.

As future work, we intend to expand the implementation of the operations to cover a more expressive family of ontologies, using the results presented in (Casanova et al., 2012a). We also intend to integrate the OntologyManagement tool with the Protégé ontology editor to take advantage of all functionalities already available in Protégé, such as ontology modeling and visualization, inference and reasoning tasks.

**Acknowledgements.** This work was partly supported by CNPq, under grants 301497/2006-0, 557128/2009-9, 483552/2009-7, 308247/2009-4, 475717/2011-2 and 552578/2011-8, by FAPERJ under grants E-26/170028/2008 and E-26/103.070/2011, by CAPES under grant NF 21/2009, and by FUNCAP under grant GPF 2151/22.

## References

Aho, A. V., Garey, M. R., Ullman, J. D., 1972. The Transitive Reduction of a Directed Graph. SIAM J. Comp. 1(2), 131–137.

Artale, A., Calvanese, D., Kontchakov, R., Zakharyaschev, M., 2009. The DL-Lite family and relations. J. of Artificial Intelligence Research 36(1), 1–69.

Baader, F., Nutt, W., 2003. Basic Description Logics. In: F. Baader, D. Calvanese, D.L. McGuiness, D. Nardi, P.F. Patel-Schneider (eds), The Description Logic Handbook: Theory, Implementation and Applications, Cambridge U. Press, Cambridge, UK, 43–95.

Berners-Lee, T., 2006. Linked Data - Design Issues. W3C.

Bizer, C., Cyganiak, R., Heath, T., 2007. How to publish Linked Data on the Web. http://www4.wiwiss.fu-berlin.de/bizer/pub/LinkedDataTutorial/

Brickley, D., Miller, L., 2010. FOAF Vocabulary Specification 0.98. Namespace Document 9. Marco Polo Edition.

Casanova, M.A., Lauschner, T., Leme, L.A.P.P., Breitman, K.K., Furtado, A.L., Vidal, V.M.P., 2010. Revising the Constraints of Lightweight Mediated Schemas. Data & Knowledge Engineering 69(12), 1274–1301.

Casanova, M.A., Breitman, K.K., Furtado, A.L., Vidal, V.M.P., Macêdo, J.A.F., 2011. The Role of Constraints in Linked Data. In: Proc. of the Confederated International Conferences: CoopIS, DOA-SVI, and ODBASE 2011, Part II. Lecture Notes in Computer Science v.7045, Springer, 781–799.

Casanova, M.A., Breitman, K.K., Furtado, A.L., Vidal, V.M.P., Macêdo, J.A.F., 2012a. An Efficient Proof Procedure for a Family of Lightweight Database Schemas. In: Michael G. Hinchey (ed.), Conquering Complexity, Springer, 431–461.

Casanova, M.A., Macêdo, J.A.F., Sacramento, E., Pinheiro, A.M.A., Vidal, V.M.P., Breitman, K.K., Furtado, A.L., 2012b. Operations over Lightweight Ontologies. In: Proc. 11th International Conference on Ontologies, DataBases, and Applications of Semantics - ODBASE 2012. Lecture Notes in Computer Science v. 7566, Springer, 646–663.

Halpin, H., Hayes, P. J., 2010. When owl:sameAs isn't the same: An analysis of identity links on the semantic web. In: Proc. International Workshop on Linked Data on the Web.

Jaffri, A., Glaser, H., Millard, I., 2008. URI disambiguation in the context of linked data. In: Proc. 1st International Workshop on Linked Data on the Web.

Jain, P., Hitzler, P., Yeh, P.Z., Verma, K., Sheth, A.P., 2010. Linked Data is Merely More Data. In: Proc. AAAI Spring Symposium: 'Linked Data Meets Artificial Intelligence', 82–86.

Hsu, H. T., 1975. An Algorithm for Finding a Minimal Equivalent Graph of a Digraph. Journal of the Association for Computing Machinery 22(1), 11–16.

Khuller, S., Raghavachari, B., Young, N., 1995. Approximating the Minimum Equivalent Digraph. SIAM Journal on Computing 24(4), 859–872.

Leme, L. A. P. P., Casanova, M. A., Breitman, K. K., Furtado, A. L., 2009. Instance-Based OWL Schema Matching. In: Proc. 11th International Conference on Enterprise Information Systems - ICEIS 2009, 14–26.

McCusker, J., McGuinness, D.L., 2010. owl:sameAs considered harmful to provenance. In: Proc. ISCB Conference on Semantics in Healthcare and Life Sciences.

Noy, N.F., Musen, M.A., 2000. PROMPT: Algorithm and Tool for Automated Ontology Merging and Alignment. In: Proc. 17th National Conference on Artificial Intelligence (AAAI-00). Available as SMI technical report SMI-2000-0831.

Noy, N.F., Kunnatur, S., Klein, M., Musen, M.A., 2004. Tracking changes during ontology evolution. In: Proc. 3rd International Semantic Web Conference, 259–273.
Pinheiro, A.M.A., 2013. "OntologyManagement Tool – Uma Ferramenta para Gerenciamento de Ontologias como Teorias Lógicas". M.Sc. Dissertation, Dept. Computing, UFC.
Prateek, J., Hitzler, P., Sheth, A.P., Verma, K., Yeh, P.Z., 2010. Ontology alignment for linked open data. In: Proc. 9th International Semantic Web Conference. Springer-Verlag, 402–417.
Raimond, Y., Giasson, F., 2010. Music Ontology Specification. Specification Document.
Ramos, J. A. Mezcla automática de ontologías y catálogos electrónicos, 2001, Final YearProject. Facultad de Informática de la Universidad Politécnica de Madrid.
Tury, M. and Bieliková, M., 2006. An Approach to Detection Ontology Changes. In: Proc. 6th International Conference on Web Engineering, 14.
Volz, R., Oberle, D. and Studer, R., 2003. Implementing Views for Light-Weight Web Ontologies. In: Proc. International Database Engineering and Application Symposium - IDEAS, 160–169.
Wang, Z., Zhang, X., Hou, L., Li, J. 2011. RiMOM2: A Flexible Ontology Matching Framework. In: Proc. ACM WebSci'11, Koblenz, Germany, 1–2.

## Appendix

**Theorem 2:** Let $\Sigma_1$ be a set of lightweight constraints and $\Sigma_2$ be the result of applying **MinimizeConstraints** to $\Sigma_1$. Then, $\Sigma_1$ and $\Sigma_2$ are equivalent, that is, $\tau[\Sigma_1]=\tau[\Sigma_2]$.

**Proof**

(a) **Minimize Graph**
Prove by induction that
(1) "⊥-node" of G($\Sigma_1$) iff "⊥-node" of $H$
(2) Condition 3 of Theorem 1 for G($\Sigma_1$) iff Condition 3 of Theorem 1 for $H$

(b) **GenerateContraints**
Prove that
(1) "⊥-node" of $H$ iff "⊥-node" of G($\Sigma_2$)
(2) Condition 3 of Theorem 1 for $H$ iff Condition 3 of Theorem 1 for G($\Sigma_2$)
Indeed, $H$ and G($\Sigma_2$) have the same arcs and the same node labels, by Proposition 2. Furthermore, by Proposition 2(vi), $\Sigma_2$ is a set of lightweight constraints.

(c) Therefore, from (a) and (b):
(1) "⊥-node" of G($\Sigma_1$) iff "⊥-node" of G($\Sigma_2$)
(2) Condition 3 of Theorem 1 for G($\Sigma_1$) iff Condition 3 of Theorem 1 for G($\Sigma_2$)
Therefore, since a "⊤-node" is the dual of a "⊥-node", by Theorem 1, we have that $\tau[\Sigma_1]=\tau[\Sigma_2]$ .□

**Theorem 3** (Correctness of **Projection**): Let $O_1 = (V_1,\Sigma_1)$ be a lightweight ontology and $W$ be a subset of $V_1$. Let $O_P = (W, \Gamma_P)$ be the ontology that **Projection** returns for $O_1$ and $W$. Then, for any lightweight inclusion $e \sqsubseteq f$ that involves only classes and properties in $W$, we have that $\Sigma_I \models e \sqsubseteq f$ iff $\Gamma_P \models e \sqsubseteq f$. □

**Proof**

Let $e \sqsubseteq f$ be a lightweight inclusion that involves only classes and properties in $W$. We have to prove that $\Sigma_I \models e \sqsubseteq f$ iff $\Gamma_P \models e \sqsubseteq f$.

($\Leftarrow$) Assume that $\Gamma_P \models e \sqsubseteq f$. Since $\Gamma_P \subseteq \tau(\Sigma_I)$ we trivially have $\Sigma_I \models e \sqsubseteq f$.

($\Rightarrow$) Assume that $\Sigma_I \models e \sqsubseteq f$. By Theorem 1, there are 3 cases to consider.

Case 1: The node of $G(\Sigma_I,\{e,f\})$ labeled with $e$ is a $\bot$-node.
Then, by Corollary 1, there is a $\bot$-node $M'$ of G($\Sigma_I$) such that $M'$ is labeled with $e'$ and $e \sqsubseteq e'$ is a tautology. Furthermore, since $e$ is a lightweight expression that involves only classes and properties in $W$ and since $e \sqsubseteq e'$ is a tautology, $e'$ is a lightweight expression that involves only classes and properties in $W$. Therefore, by Steps (3), (4), (5) of procedure **Projection** and Theorem 2, we have that $e' \sqsubseteq \bot$ is in $\Gamma_P$. Hence, since $e \sqsubseteq e'$ is a tautology, $\Gamma_P \models e \sqsubseteq \bot$. Therefore, $\Gamma_P \models e \sqsubseteq f$.

Case 2: The node of $G(\Sigma_I,\{e,f\})$ labeled with $f$ is a $\top$-node $f$.
Then, by Corollary 1, there is a $\top$-node $N'$ of G($\Sigma_I$) such that $N'$ is labeled with $f'$ and $f' \sqsubseteq f$ is a tautology. Furthermore, since $f$ is a lightweight expression that involves only classes and properties in $W$ and since $f' \sqsubseteq f$ is a tautology, $f'$ is a lightweight expression that involves only classes and properties in $W$. Then, the dual node of $N'$ is a $\bot$-node $\overline{N'}$ of G($\Sigma_I$) labeled with $\overline{f'}$. Therefore, by Steps (3), (4), (5) of procedure **Projection** and Theorem 2, we have that $\overline{f'} \sqsubseteq \bot$ is in $\Gamma_P$ and, hence, $\top \sqsubseteq f'$ is in $\Gamma_P$. Thus, since $f' \sqsubseteq f$ is a tautology, $\Gamma_P \models \top \sqsubseteq f$. Therefore, $\Gamma_P \models e \sqsubseteq f$.

Case 3: There is a path in $G(\Sigma_I,\{e,f\})$, possibly with length 0, from the node labeled with $e$ to the node labeled with $f$.
Then, by Corollary 1, there is a path in $G(\Sigma_I)$, possibly with length 0, from the node $M'$ labeled with $e'$ to the node $N'$ labeled with $f'$ such that $e \sqsubseteq e'$ and $f' \sqsubseteq f$ are tautologies. Furthermore, since $e$ is a lightweight expression that involves only classes and properties in $W$ and since $e \sqsubseteq e'$ is a tautology, $e'$ is a lightweight expression that involves only classes and properties in $W$ (and likewise for $f'$). Therefore, by Steps (3), (4), (5) of procedure **Projection** and Theorem 2, we have that $e' \sqsubseteq f'$ is in $\Gamma_P$. Thus, since $e \sqsubseteq e'$ and $f' \sqsubseteq f$ are tautology, $\Gamma_P \models e \sqsubseteq f$. □

**Theorem 4** (Correctness of **Intersection**): Let $O_1 = (V_1,\Sigma_1)$ and $O_2 = (V_2,\Sigma_2)$ be two sets of lightweight ontologies. Let $\Delta$ be the closure of $\Sigma_1$ and $\Sigma_2$ with respect to each other. Let $O_N=(V_1 \cap V_2,\Gamma_N)$ be the ontology that **Intersection** returns for $O_1$ and $O_2$. Let $e \sqsubseteq f$ be a lightweight inclusion. Then, $\Gamma_N \models e \sqsubseteq f$ iff $\Sigma_1 \models e \sqsubseteq f$ and $\Sigma_2 \models e \sqsubseteq f$.

**Proof**

Let $e \sqsubseteq f$ be a lightweight inclusion. We have to prove that $\Sigma_1 \models e \sqsubseteq f$ and $\Sigma_2 \models e \sqsubseteq f$ iff $\Gamma_N \models e \sqsubseteq f$.

(⇐) Assume that $\Gamma_N \vDash e \sqsubseteq f$. By the case analysis of Table 8 and Theorem 1, we have $\Sigma_1 \vDash e \sqsubseteq f$ and $\Sigma_2 \vDash e \sqsubseteq f$.
(⇒) Assume that $\Sigma_1 \vDash e \sqsubseteq f$ and $\Sigma_2 \vDash e \sqsubseteq f$. By Theorem 1, there are 9 cases to consider.

Case 1: For $i=1,2$, the node of $G(\Sigma_i,\{e,f\})$ labeled with $e$ is a $\bot$-node.
By Corollary 1, there is a $\bot$-node of $G(\Sigma_i)$ and, hence, of $G(\Sigma_i,\Delta)$ labeled with $g_i$ such that $e \sqsubseteq g_i$ is a tautology. Furthermore, since $e \sqsubseteq g_i$ is a tautology, we have that $g_1 \sqsubseteq g_2$ or $g_2 \sqsubseteq g_1$. Hence, there are two cases to consider: $g_1 \sqsubseteq g_2$ or $g_2 \sqsubseteq g_1$.
Case 1.1: $g_1 \sqsubseteq g_2$
Since $g_1 \sqsubseteq g_2$ and there is a $\bot$-node of $G(\Sigma_2,\Delta)$ labeled with $g_2$, there is a $\bot$-node of $G(\Sigma_2,\Delta)$ labeled with $g_1$. Hence, by case 1 of Table 8, $g_1 \sqsubseteq \bot$ is in $\Sigma_3$. Therefore, since $e \sqsubseteq g_1$ is a tautology and $g_1 \sqsubseteq \bot$ is in $\Sigma_3$, we have that $\Sigma_3 \vDash e \sqsubseteq \bot$ and, hence, $\Sigma_3 \vDash e \sqsubseteq f$.
Case 1.2: $g_2 \sqsubseteq g_1$
(Follows as in Case 1.1).

Case 2: The node of $G(\Sigma_1,\{e,f\})$ labeled with $e$ is a $\bot$-node and the node of $G(\Sigma_2,\{e,f\})$ labeled with $f$ is a $\top$-node.
By Corollary 1, there is a $\bot$-node of $G(\Sigma_1)$ and, hence, of $G(\Sigma_1,\Delta)$ labeled with $g_1$ such that $e \sqsubseteq g_1$ is a tautology. Furthermore, by Corollary 1, there is a $\top$-node of $G(\Sigma_2)$ and, hence, of $G(\Sigma_2,\Delta)$ labeled with $h_2$ such that $h_2 \sqsubseteq f$ is a tautology. Hence, by case 2 of Table 8, $g_1 \sqsubseteq h_2$ is in $\Sigma_3$. Therefore, since $e \sqsubseteq g_1$ and $h_2 \sqsubseteq f$ are tautologies and $g_1 \sqsubseteq h_2$ is in $\Sigma_3$, we have that $\Sigma_3 \vDash e \sqsubseteq f$.

Case 3: The node of $G(\Sigma_1,\{e,f\})$ labeled with $e$ is a $\bot$-node and there is a path in $G(\Sigma_2,\{e,f\})$, possibly with length 0, from the node labeled with $e$ to the node labeled with $f$.
By Corollary 1, there is a $\bot$-node of $G(\Sigma_1)$ and, hence, of $G(\Sigma_1,\Delta)$ labeled with $g_1$ such that $e \sqsubseteq g_1$ is a tautology. Furthermore, also by Corollary 1, there is a path in $G(\Sigma_2)$ and, hence, in $G(\Sigma_2,\Delta)$, possibly with length 0, from a node labeled with $g_2$ to a node labeled with $h_2$ such that $e \sqsubseteq g_2$ and $h_2 \sqsubseteq f$ are tautologies. There are 2 cases to consider: $g_1 \sqsubseteq g_2$ or $g_2 \sqsubseteq g_1$.
Case 3.1: $g_1 \sqsubseteq g_2$
Since $g_1 \sqsubseteq g_2$ and there is a path in in $G(\Sigma_2,\Delta)$, possibly with length 0, from a node labeled with $g_2$ to a node labeled with $h_2$ then there is a path in in $G(\Sigma_2,\Delta)$, possibly with length 0, from a node labeled with $g_1$ to a node labeled with $h_2$. Hence, by case 3 of Table 8, $g_1 \sqsubseteq h_2$ is in $\Sigma_3$. Therefore, since $e \sqsubseteq g_1$ and $h_2 \sqsubseteq f$ are tautologies and $g_1 \sqsubseteq h_2$ is in $\Sigma_3$, we have that $\Sigma_3 \vDash e \sqsubseteq f$.
Case 3.2: $g_2 \sqsubseteq g_1$
(Follows as in Case 3.1).

Case 4: The node of $G(\Sigma_1,\{e,f\})$ labeled with $f$ is a $\top$-node and the node of $G(\Sigma_2,\{e,f\})$ labeled with $e$ is a $\bot$-node.
(Follows as in Case 2).

Case 5: For $i=1,2$, the node of $G(\Sigma_i,\{e,f\})$ labeled with $f$ is a $\top$-node.

Then, for $i=1,2$, the node of $G(\Sigma_i,\{e,f\})$ labeled with $\bar{f}$ is a $\bot$-node. Then, applying Case 1, we have that we have that $\Sigma_3 \vDash \bar{f} \sqsubseteq \bot$ and, hence, $\Sigma_3 \vDash \top \sqsubseteq f$, which implies $\Sigma_3 \vDash e \sqsubseteq f$.

Case 6: The node of $G(\Sigma_1,\{e,f\})$ labeled with $f$ is a $\top$-node and there is a path in $G(\Sigma_2,\{e,f\})$, possibly with length 0, from the node labeled with $e$ to the node labeled with $f$.
By Corollary 1, there is a $\top$-node of $G(\Sigma_1)$ and, hence, of $G(\Sigma_1,\Delta)$ labeled with $h_1$ such that $h_1 \sqsubseteq f$ is a tautology. Furthermore, also by Corollary 1, there is a path in $G(\Sigma_2)$ and, hence, in $G(\Sigma_2,\Delta)$, possibly with length 0, from a node labeled with $g_2$ to a node labeled with $h_2$ such that $e \sqsubseteq g_2$ and $h_2 \sqsubseteq f$ are tautologies. Furthermore, since $h_i \sqsubseteq f$ is a tautology, we have that $h_1 \sqsubseteq h_2$ or $h_2 \sqsubseteq h_1$. Hence, there are two cases to consider: $h_1 \sqsubseteq h_2$ or $h_2 \sqsubseteq h_1$.
Case 6.1: $h_1 \sqsubseteq h_2$.
Since $h_1 \sqsubseteq h_2$ and there is a $\top$-node of $G(\Sigma_1,\Delta)$ labeled with $h_1$ then there is a $\top$-node of $G(\Sigma_1,\Delta)$ labeled with $h_2$. Hence, by case 6 of Table 8, $g_2 \sqsubseteq h_2$ is in $\Sigma_3$. Therefore, since $e \sqsubseteq g_2$ and $h_2 \sqsubseteq f$ are tautologies and $g_2 \sqsubseteq h_2$ is in $\Sigma_3$, we have that $\Sigma_3 \vDash e \sqsubseteq f$.
Case 6.2: $h_2 \sqsubseteq h_1$.
Since $h_2 \sqsubseteq h_1$ and there is a path in in $G(\Sigma_2,\Delta)$, possibly with length 0, from a node labeled with $g_2$ to a node labeled with $h_2$ then there is a path in in $G(\Sigma_2,\Delta)$, possibly with length 0, from a node labeled with $g_2$ to a node labeled with $h_1$. Hence, by case 6 of Table 8, $g_2 \sqsubseteq h_1$ is in $\Sigma_3$. Therefore, since $e \sqsubseteq g_2$ and $h_1 \sqsubseteq f$ are tautologies and $g_2 \sqsubseteq h_1$ is in $\Sigma_3$, we have that $\Sigma_3 \vDash e \sqsubseteq f$.

Case 7: there is a path in $G(\Sigma_1,\{e,f\})$, possibly with length 0, from the node labeled with $e$ to the node labeled with $f$ and the node of $G(\Sigma_2,\{e,f\})$ labeled with $e$ is a $\bot$-node.
(Follows as in Case 3).

Case 8: there is a path in $G(\Sigma_1,\{e,f\})$, possibly with length 0, from the node labeled with $e$ to the node labeled with $f$ and the node of $G(\Sigma_1,\{e,f\})$ labeled with $f$ is a $\top$-node.
(Follows as in Case 6).

Case 9: For $i=1,2$, there is a path in $G(\Sigma_i,\{e,f\})$, possibly with length 0, from the node labeled with $e$ to the node labeled with $f$.
Then, by Corollary 1, there is a path in $G(\Sigma_i)$ and, hence, in $G(\Sigma_i,\Delta)$, possibly with length 0, from a node labeled with $g_i$ to a node labeled with $h_i$ such that $e \sqsubseteq g_i$ and $h_i \sqsubseteq f$ are tautologies. There are 4 cases to consider, generated by a combination of the assumptions: ($g_1 \sqsubseteq g_2$ or $g_2 \sqsubseteq g_1$) and ($h_2 \sqsubseteq h_1$ or $h_1 \sqsubseteq h_2$).
Case 9.1: $g_1 \sqsubseteq g_2$ and $h_2 \sqsubseteq h_1$.
Then, for $i=1,2$, there is a path in $G(\Sigma_i,\Delta)$, possibly with length 0, from a node labeled with $g_1$ to a node labeled with $h_1$. By Case 9 of Table 8, $g_1 \sqsubseteq h_1$ is in $\Sigma_3$. Since $e \sqsubseteq g_1$ and $h_1 \sqsubseteq f$ are tautologies, we have that $\Sigma_3 \vDash e \sqsubseteq f$.
Case 9.2: $g_2 \sqsubseteq g_1$ and $h_1 \sqsubseteq h_2$.
(Follows as in Case 3.1).
Case 9.3: $g_1 \sqsubseteq g_2$ and $h_1 \sqsubseteq h_2$.
Then, for $i=1,2$, there is a path in $G(\Sigma_i,\Delta)$, possibly with length 0, from a node labeled with $g_1$ to a node labeled with $h_2$. By Case 9 of Table 8, $g_1 \sqsubseteq h_2$ is in $\Sigma_3$. Since $e \sqsubseteq g_1$ and $h_2 \sqsubseteq f$ are tautologies, we have that $\Sigma_3 \vDash e \sqsubseteq f$.

Case 9.4: $g_2 \sqsubseteq g_1$ and $h_2 \sqsubseteq h_1$.
(Follows as in Case 3.3). □

**Theorem 5** (Correctness of **Difference**): Let $O_1 = (V_1, \Sigma_1)$ and $O_2 = (V_2, \Sigma_2)$ be two sets of lightweight ontologies. Let $\Delta$ be the closure of $\Sigma_1$ and $\Sigma_2$ with respect to each other. Let $O_D = (V_1, \Gamma_D)$ be the ontology that **Difference** returns for $O_1$ and $O_2$. Then, for any lightweight inclusion $e \sqsubseteq f$ such that $\Gamma_D \models e \sqsubseteq f$, we have that $\Sigma_1 \models e \sqsubseteq f$ but not $\Sigma_2 \models e \sqsubseteq f$.

**Proof**

Let $e \sqsubseteq f$ be a lightweight inclusion. Assume that $\Gamma_D \models e \sqsubseteq f$. By the case analysis of Table 10 and Theorem 1, we have $\Sigma_1 \models e \sqsubseteq f$ but not $\Sigma_2 \models e \sqsubseteq f$. □